\documentclass[twoside,11pt]{article}

\usepackage{jmlr2e}

\newcommand{\E}{\mathbb{E}}
\newcommand{\pluseq}{\mathrel{+}=}
\newcommand{\minuseq}{\mathrel{-}=}
\newcommand{\g}{\, | \,}
\newcommand{\R}{\mathbb{R}}

\usepackage{amsmath}
\usepackage{booktabs}
\usepackage{lscape}

\usepackage[nameinlink,noabbrev]{cleveref}
\crefname{appsec}{Appendix}{Appendices}
\creflabelformat{equation}{#2\textup{#1}#3}
\crefname{algocf}{algorithm}{algorithms}

\usepackage[usenames,dvipsnames]{xcolor}
\definecolor{shadecolor}{gray}{0.9}
\definecolor{Green}{HTML}{156946}

\usepackage{graphicx}
\usepackage[format=hang]{subcaption}
\usepackage{rotating}

\usepackage{booktabs}

\usepackage[algoruled]{algorithm2e}
\usepackage{listings}
\usepackage{fancyvrb}
\fvset{fontsize=\normalsize}

\usepackage{times} 
\usepackage{amsmath}

\renewcommand*\copyright{{\textcopyright}}

\ShortHeadings{Nonparametric Deconvolution Models}{Chaney, Verma, Lee, \& Engelhardt}
\firstpageno{1}

\begin{document}

\title{Nonparametric Deconvolution Models}

\author{\name Allison J.B. Chaney \email ajb.chaney@duke.edu \\
       \addr Fuqua School of Business\\
       100 Fuqua Drive\\
       Duke University\\
       Durham, NC 27708, USA
       \AND
       \name Archit Verma \email architv@princeton.edu \\
       \addr Department of Chemical and Biological Engineering\\
       Princeton University\\
       Princeton, NJ 08540, USA
       \AND
       \name Young-suk Lee \email youngl@cs.princeton.edu \\
       \addr Department of Computer Science\\
       Princeton University\\
       Princeton, NJ 08540, USA
       \AND
       \name Barbara E. Engelhardt \email bee@princeton.edu \\
       \addr Department of Computer Science\\
       \addr Center for Statistics and Machine Learning\\
       Princeton University\\
       Princeton, NJ 08540, USA}

\editor{}

\maketitle

\begin{abstract}
We describe \emph{nonparametric deconvolution models} (NDMs),
a family of Bayesian nonparametric models for collections of data in which each observation is the average over the features from heterogeneous particles. For example, these types of data are found in elections, where we observe precinct-level vote tallies (observations) of individual citizens' votes (particles) across each of the candidates or ballot measures (features), where each voter is part of a specific voter cohort or demographic (factor).  Like the hierarchical Dirichlet process, NDMs rely on two tiers of Dirichlet processes to explain the data with an unknown number of latent factors; each observation is modeled as a weighted average of these latent factors.  Unlike existing models, NDMs recover how factor distributions vary locally for each observation.
This uniquely allows NDMs both to deconvolve each observation into its constituent factors, and also to describe how the factor distributions specific to each observation vary across observations and deviate from the corresponding global factors.
We present variational inference techniques for this family of models and study its performance on simulated data and voting data from California.
We show that including local factors improves estimates of global factors and provides a novel scaffold for exploring data.\looseness=-1
\end{abstract}


\section{Introduction}
Consider a collection of citizens in a voting precinct.  Each voter cast their votes for candidates and issues, and the votes are aggregated together into the precinct-level vote.  These data may be analyzed for many purposes: forecasting, campaign targeting, developing community programs, and understanding the composition of the electorate.  For all of these applications, it is useful to identify voting cohorts---groups of people that vote similarly and often share population demographics such as gender, socioeconomic status, or race. Then one may study these voter cohorts in terms of voter cohort prevalence within each precinct and how the cohort voted across precincts.

Several model families exist to decompose observations such as these into patterns that can be interpreted as cohorts.  However, no existing model family captures the notion that the distribution of a voting cohort may vary locally within precincts.  For example, middle-class voters in a precinct with a recent incidence of gun violence might systematically vote more in favor of gun control than the middle-class cohort counterparts in other precincts.  In other words, it is important to know how voters of a given cohort voted \emph{within a specific precinct}, and, in turn, how the precinct-specific cohort votes differ from the global cohort votes.

In this paper, we introduce \emph{deconvolution models}, a new class of mixed membership models with observation-specific mixture distributions.  This model family is designed for data with \emph{convolved} observations such as ballot outcomes---each observation (e.g., outcomes for a voting precinct) is composed of many particles (e.g., voters casting their votes); these particles are observed in aggregate, or convovled together into an observation with some number of features (e.g., measures on a ballot).  This same structure exists in data from many disciplines, including politics, sports, finance, and biology, as shown in \Cref{tab:nomenclature}. The goal of deconvolution models is to explicitly model observation-specific deviations from the global factor distributions.\looseness=-1

The term \emph{deconvolution} is used in many settings to denote similar notions of decomposing information, and often carries specific technical connotations in different contexts.  In signal processing, it is used to refer to two conceptually distinct processes; its use in density deconvolution and convolutional neural networks further adds to the confusion.  In all of these settings, the core meaning behind the term is the same: some value is a convolution, or blending, of component parts; deconvolution involves estimating these unknown components.  
Moving forward, we use the term \emph{deconvolution} to refer to estimates of both the local (observation-specific) and global factor distributions and the proportions of those factors represented in each observation; we build models to allow us to estimate those three components essential to deconvolution.\looseness=-1

This paper is organized as follows. We first place this work in the context of related models in \Cref{sec:related_work}.
In \Cref{sec:model_family}, we formally define the nonparametric deconvolution model (NDM) family and describe several instances of this family in \Cref{sec:model_instances}.
Then, we describe an inference algorithm\footnote{Open source software for our inference algorithm is available at \url{https://github.com/ajbc/ndms}.} for estimating the posterior of these models in \Cref{sec:inference}. We explore the resulting approximations and compare results with related methods on simulated data and California voting data in \Cref{sec:results}.  We conclude with a discussion of the advantages, limitations, and potential extensions of this model family in \Cref{sec:conclusion}.

\begin{table}[!ht]
\small
\centering
\begin{tabular}{ccccccc}
\toprule
\textbf{General} & \textbf{Voting} & \textbf{Bulk RNA-seq} & \textbf{fMRI} & \textbf{Baseball} \\
\midrule
observation $y_n$ & precinct votes & sample expression levels & image & pitcher \\
feature $m$ & issue or candidate & gene & voxel & pitch type and outcome \\
particle $p$ & individual voter & one cell & one neuron & one pitch \\
factor $k$ & voting cohort & cell type & response pattern & pitching strategy \\
\bottomrule
\end{tabular}
\caption{Structure of convolved observations in multiple domains.}
\label{tab:nomenclature}
\end{table}

\section{Related Work on Statistical Deconvolution}
\label{sec:related_work}
\begin{figure}
\centering
\includegraphics[width=\textwidth]{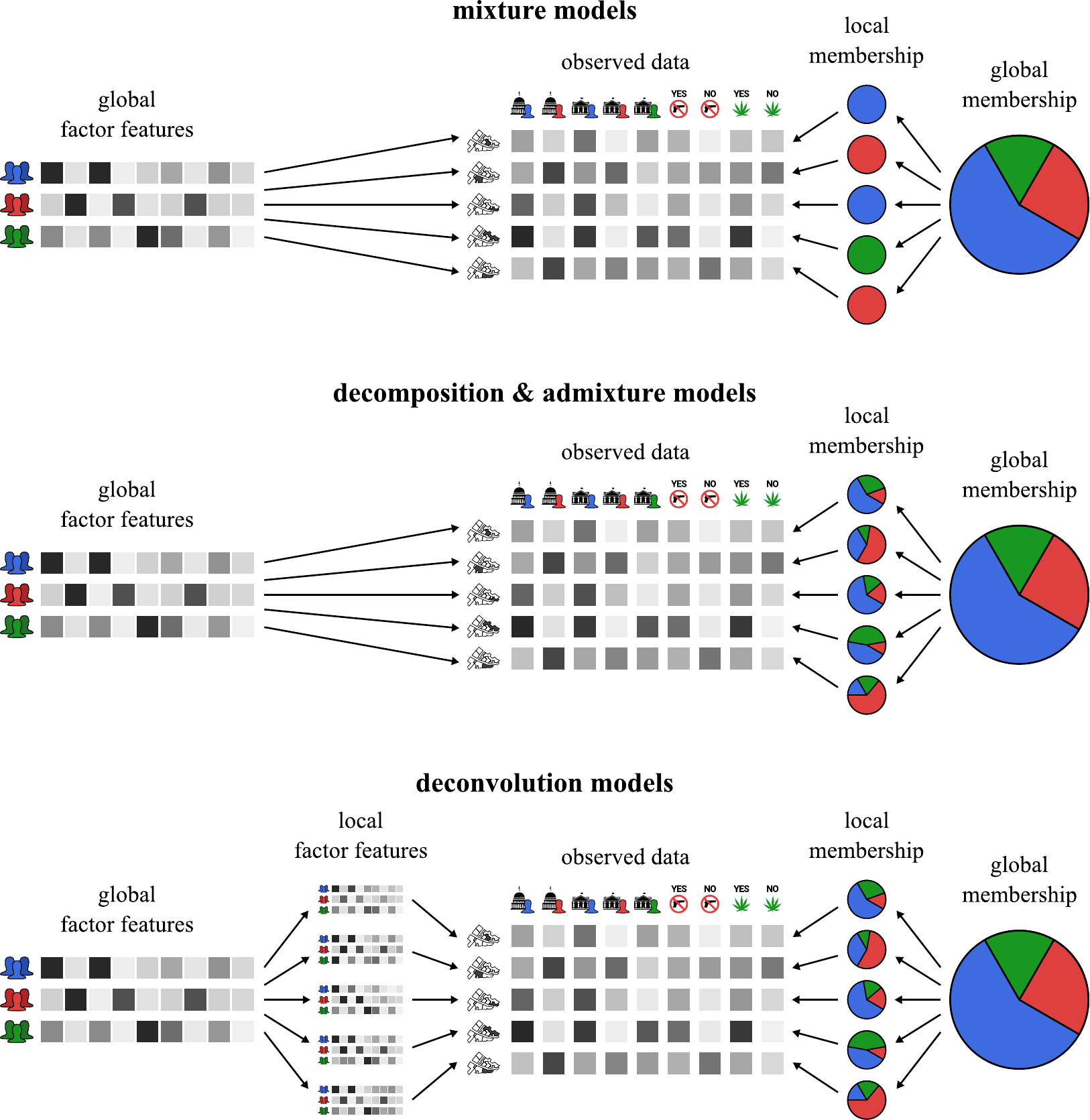}
\vspace{10px}
\caption{Illustrations of multiple latent variable models.  
\emph{Mixture models} assign each observation to one of $K$ clusters, or factors.  
\emph{Decomposition} and \emph{admixture models} both model observations with local factor membership proportions.
\emph{Deconvolution models} (this paper) also include observation-specific (local) factor feature distributions.}
\label{fig:cartoon}
\end{figure}

Many disciplines rely on the analysis of high-dimensional heterogeneous data; latent variable models are well-suited to expose hidden patterns in these data.  The simplest form of the latent variable model is a \emph{mixture model} (\Cref{fig:cartoon}, top), which assign each observation to one of $K$~clusters.  For each cluster~$k$, there is an associated probability distribution on each feature---when an observation is assigned to cluster $k$, it is assumed to be drawn from the corresponding cluster distribution. Whether the observations have hard (i.e., single cluster) assignments or soft (i.e., probabilistic) assignments, the generative model assumes that each observation comes from only a single cluster distribution. Global membership then captures the proportion of observations assigned to each cluster.\looseness=-1

More complex latent factor models build on this structure, relying on similar notions of global membership and factor-specific feature distributions. For example, \emph{admixture models} are the simplest version of a mixed-membership model. In admixture models, each observation is generated from a convex combination (i.e., a weighted sum where the weights are non-negative and sum to one) of the $K$ latent factor distributions~\citep{pritchard2000inference}; this is in contrast to the observations being generated from a single factor's distribution, as in the mixture model framework.  These latent factor distributions, as in the mixture model, are shared across all observations. Latent Dirichlet allocation \citep[LDA;][]{Blei:2003b} is a well-known instance of this model family where the global membership variables have Dirichlet priors.  The hierarchical Dirichlet process~\citep[HDP;][]{Teh:2006} extends LDA with a Bayesian nonparametric prior that enables support over an infinite number of latent factors and the ability to share feature distributions across multiple, nested collections of data.\looseness=-1

Admixture models are a subset of the broader class of \emph{decomposition models} (\Cref{fig:cartoon}, middle), which represent local factor membership as a mixture of global factors.
Matrix factorization is a popular model family and comes in many varieties, including non-negative matrix factorization~\citep[NMF;][]{Lee01};\footnote{Note that the original NMF construction is not a probabilistic generative model and no likelihood or  associated posterior distribution is available; this posterior distribution is important when interested in variance.} Gamma-Poisson matrix factorization~\citep[GaP;][]{CannyGaP}, and Gaussian probabilistic matrix factorization~\citep[PMF;][]{PMF}.
Other examples of decomposition models include principal component analysis~\citep[PCA;][]{hotelling1933analysis,jolliffe1986principal, tipping1999probabilistic}, factor analysis~\citep[FA;][]{harman1960modern,jolliffe1986principal}, and
their sparse variants~\citep[respectively]{zou2006sparse,engelhardt2010analysis}.

We use the term \emph{deconvolution models} to distinguish a new class of mixed membership models with observation-specific factor distributions across features.  While \emph{deconvolution} is used to refer to a variety of concepts, we use it here to refer to a family of mixed membership models that include local (i.e., observation-specific) factor feature distributions (\Cref{fig:cartoon}, bottom).  Specifically, deconvolution models draw on decomposition models for the notion of group-specific distributions of membership and global factor features shared among all observations.  But unlike these models, deconvolution introduces observation-specific (or local) factor feature distributions to capture real-world variation in the factor feature distributions.
This model structure is most advantageous in the context of \emph{convolved admixed data}, where observations are collections of heterogeneous particles that have been averaged or otherwise convolved together to be observed as a single unit; real-world examples of convolved data are shown in \Cref{tab:nomenclature}.  The objective of a deconvolution model is to reverse this convolution process in order to both estimate the factor proportions of the underlying particles in each observation, as well as to estimate the feature values of all particles assigned to a specific factor within an observation; \Cref{sec:model_family} will describe this model family in greater detail. 

\section{Nonparametric Deconvolution Models}
\label{sec:model_family}
In this section, we formally specify the family of nonparametric deconvolution models (NDMs).  We begin by describing the parametric variant of this model family, which includes a fixed number of latent factors $K$ (\Cref{sec:parametric_model_family}).  The nonparametric version of this model family (which estimates $K$) is based on the hierarchical Dirichlet process \citep[HDP,][]{Teh:2006}, a Bayesian nonparametric admixture model. We will review the normalized gamma process construction of the HDP~\citep{Paisley:2012} in \Cref{sec:hdp}. Then, we will introduce the NDM family (\Cref{sec:generativeProcess}) and describe several instances of this family (\Cref{sec:model_instances}).

\subsection{Parametric Deconvolution Models}
\label{sec:parametric_model_family}

The parametric variant of the proposed deconvolution model family requires a fixed number of latent factors $K$.  Each factor $k$ is present in the global population with proportion $\beta_k$; a randomly chosen particle (e.g., an individual voter) has probability $\beta_k$ of being associated with factor $k$ (e.g., one voting cohort).  We assume that these global factor proportions are drawn from a Dirichlet distribution parameterized by $\boldsymbol{\alpha}_0$,
\begin{equation}
\boldsymbol{\beta} \g \boldsymbol{\alpha}_0 \sim \mbox{Dirichlet}(\boldsymbol{\alpha}_0). \label{eq:parametric_beta}
\end{equation}

Similarly, we assume each of the $n$ convolved observations has observation-specific (or local) proportions $\boldsymbol{\pi}_n$, where $\boldsymbol{\pi}_{n,k}$ represents the probability that a random particle from observation $n$ (e.g., a voter from Alameda County) will be associated with factor $k$.  As with the global proportions, we assume these local proportions are drawn from a Dirichlet distribution; the distribution is parameterized using the global proportions $\boldsymbol{\beta}$ scaled by hyperparameter $\alpha$,
\begin{equation}
\boldsymbol{\pi}_n \g \boldsymbol{\beta}, \alpha \sim \mbox{Dirichlet}(\alpha \boldsymbol{\beta}).
\label{eq:parametric_pi}
\end{equation}

To describe the feature distribution of each latent factor $k$, we use a combination of two parameters at the global level: mean $\boldsymbol{\mu}_k \in \R^{M}$ and covariance matrix $\boldsymbol{\Sigma}_k \in \R^{M\times M}$.  Each global mean $\boldsymbol{\mu}_k$ represents the average value of each of $M$ features over all particles from factor $k$; the covariance matrix $\boldsymbol{\Sigma}_k$ represents the covariance of these $M$ features among particles from factor $k$.  The latent factor feature parameters $\boldsymbol{\phi} $ are the set of these two parameters, $\boldsymbol{\phi}=\{\boldsymbol{\mu}, \boldsymbol{\Sigma}\}$.
We assume that the global mean $\mu_{k,m}$ for factor $k$ and feature $m$ is drawn from a normal distribution,
\begin{equation}
\mu_{k,m} \g \mu_0, \sigma_0 \sim \mathcal{N}(\mu_0, \sigma_0), \label{eq:parametric_mu}
\end{equation}
and that covariance matrix $\boldsymbol{\Sigma}_k$ is drawn from an inverse Wishart distribution,
\begin{equation}\boldsymbol{\Sigma}_{k} \g \nu, \boldsymbol{\Psi} \sim \mathcal{W}^{-1}(\boldsymbol{\Psi}, \nu). \label{eq:parametric_Sigma}\end{equation}

In an ideal world, we would know the number of particles $P_n$ associated with observation $n$.  If we were given this information, we could model the assignments $z_{n,p}$ of each particle $p$ to the $K$ factors,
\begin{equation}
z_{n,p} \g \boldsymbol{\pi}_n \sim \mbox{Categorical}\left(\boldsymbol{\pi}_n\right), \label{eq:parametric_x_assignment}
\end{equation}
and then draw the local particle-specific features from the global features associated with its assigned factor $z_{n,p}$,
\begin{equation}
\boldsymbol{x}_{n,p} \g z_{n,p}, \boldsymbol{\mu}, \boldsymbol{\Sigma} \sim \mathcal{N}_{M}(\boldsymbol{\mu}_{z_{n,p}}, \boldsymbol{\Sigma}_{z_{n,p}}).
\label{eq:parametric_x_value}
\end{equation}

In practice, however, we do not need to infer the values for the particle assignments $\boldsymbol{z}$ and particle-specific features $\boldsymbol{x}$.  Instead of modeling the $K$ features of each particle $p$ with $x_{n,p,k}$, we model the average of these particles for a given factor, or 
\begin{equation}
\boldsymbol{\bar{x}}_{n,k} = \frac{1}{\sum_{p=1}^{P_n}\mathbf{1}(z_{n,p}=k)} \sum_{p=1}^{P_n}\mathbf{1}(z_{n,p}=k)~\boldsymbol{x}_{n,p}.
\label{eq:parametric_x_ave}
\end{equation}
Thus, just as we model latent factor proportions at the local, or observation-specific, level with $\boldsymbol{\pi}_n$, we use variables $\boldsymbol{\bar{x}}_{n,k}$ to describe the latent feature values for   all the particles in observation $n$ associated with factor $k$. Using the fact that the sum of normally-distributed variables is also normally-distributed, we assume these averaged local features $\boldsymbol{\bar{x}}$ are drawn from $M$-dimensional multivariate normal distributions,
\begin{equation}
\boldsymbol{\bar{x}}_{n,k} \g \boldsymbol{\pi}_{n,k}, \boldsymbol{\mu}_{k}, \boldsymbol{\Sigma}_{k} \sim \mathcal{N}_M \left(\boldsymbol{\mu}_k,~ \frac{\boldsymbol{\Sigma}_k}{P_n \pi_{n,k}}\right).
\label{eq:parametric_x}
\end{equation}
This construction allows us to study local features without needing to infer particle assignments $\boldsymbol{z}_{n,p}$ or particle-specific feature values $\boldsymbol{x}_{n,p}$. These variables capture observation-specific deviations from the global factor distributions; they allow us to answer questions such as ``how do voters from a specific cohort vote in \emph{this particular precinct}?''

Practically, we still need the number of particles $P_n$ for observation $n$, which is often not available, but an approximation.  Thus, we explicitly model the number of particles $P_n$ when needed, or
\begin{equation}
P_n \g \rho \sim \mbox{Poisson}(\rho).
\label{eq:parametric_p}
\end{equation}
Parameter $\rho$ can be set according to a rough estimate of $P_n$.

To complete the parametric model specification, we provide a framework for generating observations $y_{n,m}$ for observation $n$ (e.g., a voting precinct) and feature $m$ (e.g., an issue or candidate).  We combine the local features $\boldsymbol{\bar{x}_{n,k}}$ of observation $n$ over all $K$ latent factors, then we pass this weighted average through a link function $g$ and use this result to parameterize some distribution $f$.  Broken down for a single feature $m$, we generate an observation $n$
\begin{equation}
y_{n,m} \g \boldsymbol{\bar{x}}_{n}, \boldsymbol{\pi}_n \sim f\left(g\left(\sum_{k=1}^K \pi_{n,k}~ \bar{x}_{n,k,m}\right)\right). \label{eq:parametric_y}
\end{equation}
Though the generative processes for a nonparametric deconvolution model (\Cref{sec:generativeProcess}) is different from this parametric generative process, they both correspond to the same graphical model (\Cref{fig:graphical_model}), except the parametric version uses $K$ latent factors instead of an infinite number.

While we will discuss instances of the full model family in detail within \Cref{sec:model_instances}, we will briefly preview some example model instances here.  As with generalized linear models (GLMs), if we choose the distribution $f$ to be Gaussian, the link function $g$ is most naturally the identity function.  Similarly, if we think the observations $\boldsymbol{y}$ are best modeled using a Poisson distribution for $f$, the canonical link function $g$ would be an exponential in order to transform the combination of local Gaussian features $\sum_k \pi_{n,k} \boldsymbol{\bar{x}}_{n,k}$ to be positive real values to parameterize the Poisson distribution.

Some model families may require additional feature-specific hyperparameters---for example, when $f$ is Gaussian, we may want feature-specific variances.  We refer to feature-specific hyperparameters as $\eta_m$ for feature $m$; in practice we set them to be relatively small values so that the majority of variance is captured by the estimated parameters.
If additional hyperparameters are needed, we may adjust \Cref{eq:parametric_y} to include them:
\begin{equation}
y_{n,m} \g \boldsymbol{\bar{x}}_{n}, \boldsymbol{\pi}_n \boldsymbol{\eta} \sim f\left(g\left(\sum_{k=1}^K \pi_{n,k}~ \bar{x}_{n,k,m} \right), \eta_m\right). \label{eq:parametric_y_with_eta}
\end{equation}

\subsection{Background: construction of the HDP}
\label{sec:hdp}
We will now extend this preliminary parametric decomposition model family to its full nonparametric form.  We first review the hierarchical Dirichlet process (HDP), upon which we base the nonparametric version of the deconvolution model family.

The HDP \citep{Teh:2006} is constructed using two layers of Dirichlet processes (DPs).  This hierarchical process defines a global random probability measure $G_0$ and a set of random probability measures $G_n$, one for each group $n$. In our case, these measures specify both factor proportions and feature distributions.  The global measure $G_0$ is distributed as a Dirichlet process with concentration parameter $\alpha_0$ and base probability measure $H$: 
\begin{equation}
G_0 \g \alpha_0, H \sim \mbox{DP}(\alpha_0, H).
\end{equation}
The random measures $G_n$ are also distributed as Dirichlet processes with base measure $G_0$:
\begin{equation}
G_n \g \alpha, G_0 \sim \mbox{DP}(\alpha, G_0).
\end{equation}
A hierarchical Dirichlet process can be used as the prior distribution over the factors for grouped data, allowing us to define hierarchical Dirichlet process mixture models.  A single Dirichlet process can be constructed in several equivalent ways, the choice of construction has implications for inference.

Following~\citet{Paisley:2012}, we use two different representations of the DP---one for each layer. The top-level DP uses the~\citet{Sethuraman:1994} stick-breaking representation of the DP.  As its name suggests, we imagine that some population may be partitioned into its component parts much the way one would break a stick into pieces.  Formally, we represent global proportions as $\beta_k$; this could describe, for example, how many middle class people there are as a percentage of all voters.  The Sethuraman generative process draws the unnormalized variant of these proportions $\beta_k'$ from a beta distribution:
\begin{equation}
\beta'_k \g \alpha_0 \sim \mbox{Beta}(1,\alpha_0). 
\label{eq:unormalizedBeta}
\end{equation}
The hyperparameter $\alpha_0$ is called the \emph{concentration parameter} and controls the distribution over the proportions, with higher values leading to smaller numbers of larger clusters and values closer to $0$ leading to larger numbers of smaller clusters in expectation.
These proportions are normalized relative to all previous proportions,
\begin{equation}
\beta_k = \beta'_k \prod_{\ell=1}^{k-1} \left(1-\beta'_{\ell}\right), \label{eq:normalizedBeta}
\end{equation}
which gives us the stick breaking analogy: starting with a stick of length $1$, after partitioning some portion of the population into $k-1$ factors (breaking the stick $k-1$ times), for the remainder of the population (or stick), what proportion should I assign to factor $k$ (or how much of the stick should I break off)?

With $\beta_k$ as our global proportions for factor $k$ (e.g., what proportion of voters belong to voting cohort $k$?), we represent the distribution of features associated with factor $k$ as $\boldsymbol{\phi}_k$ (e.g., how do people in cohort $k$ vote across all districts?).  We generate these parameters from the base distribution $H$:\looseness=-1
\begin{equation}
\boldsymbol{\phi}_k \g H \sim H.
\label{eq:phi}
\end{equation}
When the HDP is used for modeling multinomial features, as topic models capture bag-of-words representations of text, the base distribution is usually a symmetric Dirichlet distribution over the feature simplex.  For NDMs, $H$ will take an alternative form to align with the parametric deconvolution model generative process (\Cref{eq:parametric_mu,eq:parametric_Sigma}).

The second layer of the HDP captures the relationship between the local level and the global level.
The local proportions $\pi_{n,k}$ are the analog of the global proportions $\beta_k$, with one set of proportions for each observation $n$; these are the observation-specific factor proportions. In the context of voting precincts, $\pi_{n,k}$ tells us what proportion of precinct $n$ is made up of voting cohort $k$.
To generate these local proportions, we use a normalized gamma process prepresentation of the DP~\citep{Ferguson:1973, Paisley:2012}, which begins by generating unnormalized proportions from a gamma distribution,
\begin{equation}
\pi'_{n,k} \g \alpha, \beta_k \sim \mbox{Gamma}(\alpha \beta_k, 1), 
\label{eq:unnormalizedPi}
\end{equation}
and then normalizes them:
\begin{equation}
\pi_{n,k} = \frac {\pi'_{n,k}} {\sum_{\ell=1}^{\infty} \pi'_{n, \ell}}. 
\label{eq:normalizedPi}
\end{equation}
These normalized proportions $\pi_{n,k}$ are Dirichlet-distributed because of the relationship between the gamma and Dirichlet distributions \cite{Ferguson:1973}.
Like $\alpha_0$, hyperparameter $\alpha$ is a concentration parameter encoding the distance of the local proportions from the global proportions. 

\begin{figure}[tb]
    \begin{minipage}[c]{0.40\linewidth}
        \centering
        \includegraphics[height=120px]{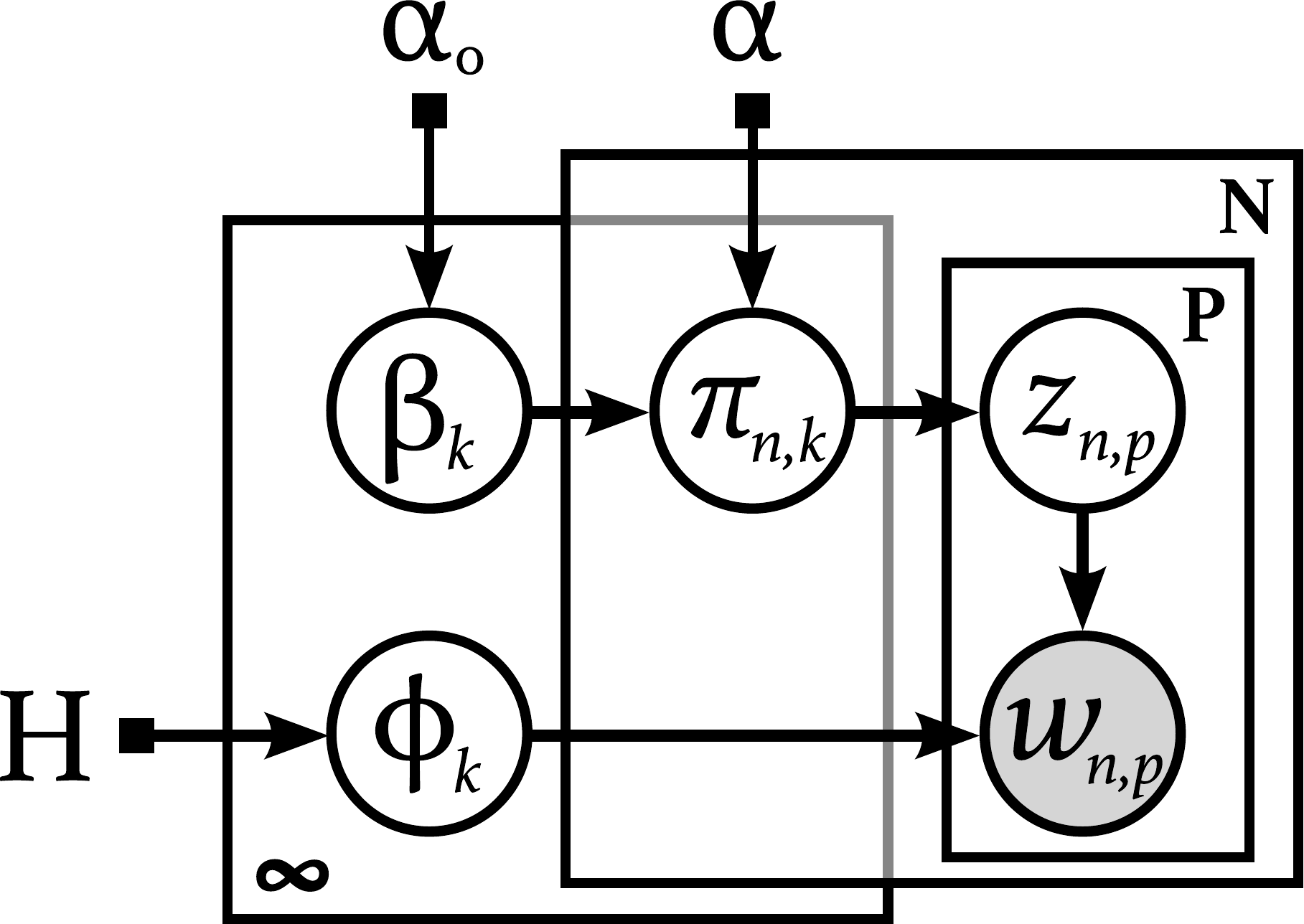}
    \end{minipage}
    \hspace{0.5cm}
    \begin{minipage}[t]{0.50\linewidth}
        \footnotesize
        \begin{tabular}{cl}
        \toprule
        & \textbf{Notation} \\
        \midrule
         $\infty$ & number of latent topics \\
         $N$ & number of documents \\
         $P$ & number of words in a document \\
         $\boldsymbol{\beta}$ & global distribution of topics  \\
         $\boldsymbol{\pi}$ & per-document (local) distribution of topics \\
         $\boldsymbol{\phi}$ & topics (one distribution over vocabulary terms per topic) \\
         $\boldsymbol{z}$ & per-word topic assignments \\
         $\boldsymbol{w}$ & observed words \\
         \bottomrule
        \end{tabular}
    \end{minipage}
    \caption{Graphical model for a hierarchical Dirichlet process mixture model~\citep{Teh:2006} with corresponding notation, framed in the topic modeling context.}
    \label{fig:hdp}
\end{figure}

The NDM family uses this same construction but, at this point, the models diverge.  HDP mixture models 
(e.g., \Cref{fig:hdp}) continue by using the local factor proportions $\boldsymbol{\pi}$ and the global factor features $\boldsymbol{\phi}$ to construct discrete probability distributions $G_n$, one for each group of observations $n$:
\begin{equation}
G_n = \sum_{k=1}^\infty \pi_{n,k} \delta_{\phi_k}.
\end{equation}
Each $G_n$ is a distribution over the global factors $\phi_k$---a draw from $G_n$ produces $\phi_k$ with probability $\pi_{n,k}$.\footnote{Other constructions simply draw an index to factor $k$ with probability $\pi_{n,k}$.  Either way, $z_{n,k}$ depends on both $\pi_{n,k}$ and $\phi_k$---this construction just requires less bookkeeping.}
Thus we can use $G_n$ to draw local factor assignments $z_{n,p}$ for particle $p$ in group $n$,
\begin{equation}z_{n,p} \g G_n \sim G_n.\end{equation}
In the voting example, $z_{n,p}$ is the voting cohort assigned to voter $p$ of precinct $n$.  Notably, the votes of voter $p$ are assumed to be observed in the HDP mixture model setting; the observed data $w_{n,p}$ (e.g., the the votes cast by voter $p$ in precinct $n$) are then drawn from a distribution $F$ parameterized by $z_{n,p}$:
\begin{equation}
w_{n,p} \g z_{n,p} \sim F(z_{n,p});
\end{equation}
as the prior distribution $H$ must be conjugate to $F$ for the inference algorithm updates to be closed form, $F$ is usually multinomial.

The individual factor assignments $z_{n,p}$ and observations $w_{n,p}$ are marginalized out in the NDM family, as convolved admixed data does not involve observations of individual particles---for instance, we only record votes aggregated at the precinct level for privacy reasons.

\subsection{NDM Generative Process}
\label{sec:generativeProcess}
As with the HDP construction, the NDM family draws global factor proportions $\beta'_k$ (\Cref{eq:unormalizedBeta}) and normalizes them to $\beta_k$ (\Cref{eq:normalizedBeta}, ``how much of the population is in voting cohort $k$?'').  
We likewise represent global factor feature distributions---``how do people in cohort $k$ usually vote?''---but instead of using the general form of $\boldsymbol{\phi}_k$~(\Cref{eq:phi}), we describe the features for each global factor in terms of its mean $\boldsymbol{\mu}_k$ (\Cref{eq:parametric_mu}) and covariance matrix $\boldsymbol{\Sigma}_k$ (\Cref{eq:parametric_Sigma}).  This still follows the general HDP framework (\Cref{eq:phi}), and can be viewed as $\boldsymbol{\phi}_k=\{\boldsymbol{\mu}_k, \boldsymbol{\Sigma}_k\}$.

At the local level, we also have factor proportions $\pi'_{n,k}$ (\Cref{eq:unnormalizedPi}) that are similarly normalized to $\pi_{n,k}$ (\Cref{eq:normalizedPi}, ``How much of this precinct is in the middle class cohort?'').  Deviating from the HDP construction, we draw \emph{local factor features} $\boldsymbol{\bar{x}}_{n,k}$ (\Cref{eq:parametric_x}), which enable us to identify how local feature distributions of each factor $k$ deviate from global ones.  With the voting example, instead of assuming that the middle class votes the same in every precinct, we can characterize how the middle class votes in each precinct separately---one precinct may vote more socially liberal relative to global patterns.

\begin{figure}[tb]
    \begin{minipage}[c]{0.45\linewidth}
        \centering
        \includegraphics[height=150px]{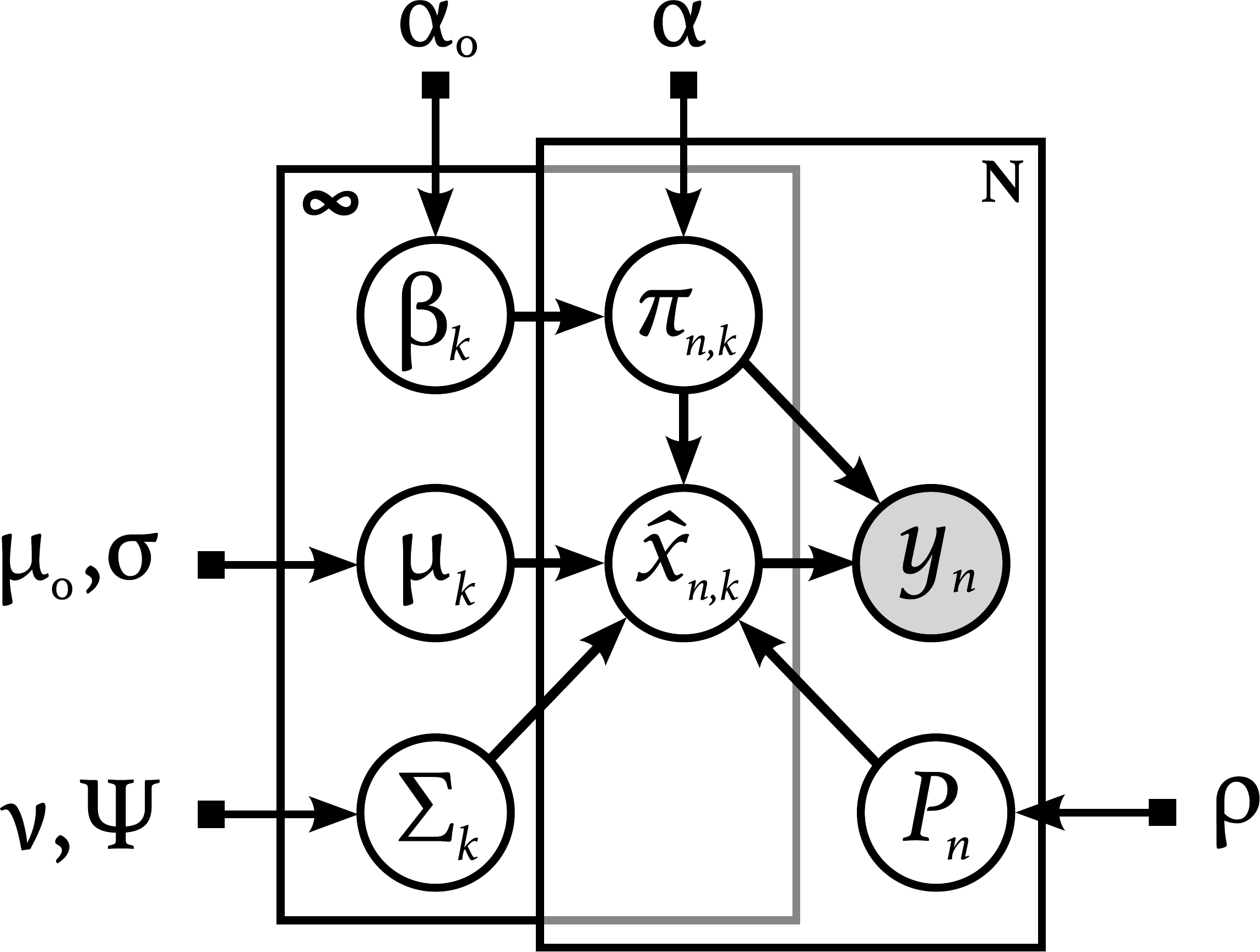}
    \end{minipage}
    \hspace{0.5cm}
    \begin{minipage}[t]{0.45\linewidth}
        \footnotesize
        \begin{tabular}{cl}
        \toprule
        & \textbf{Notation} \\
        \midrule
         $\infty$ & number of latent factors \\
         $N$ & number of observations \\
         $M$ & number of features for each factor or observation \\
         $\boldsymbol{\beta}$ & global factor proportions ($\infty$-dimensional vector)\\
         $\boldsymbol{\pi}$ & local factor proportions ($N \times \infty$) \\
         $\boldsymbol{\mu}$ & global factor feature means ($\infty \times M$) \\
         $\boldsymbol{\Sigma}$ & global factor feature covariances ($\infty \times M \times M$) \\
         $\boldsymbol{\bar{x}}$ & local factor features ($N \times \infty \times M$) \\
         $\boldsymbol{P}$ & local number of particles ($N$) \\
         $\boldsymbol{y}$ & convolved observations ($N \times M$) \\
         \bottomrule
        \end{tabular}
    \end{minipage}
   \caption{Graphical model for nonparametric deconvolution models (NDMs) with corresponding notation.}
\label{fig:graphical_model}
\end{figure}

At the local level, we also draw the number of particles $P_n$ for each observation $n$ (\Cref{eq:parametric_p}).
Given the local parameters, we then generate our observations $\boldsymbol{y}_n$ just as in the parametric model variant (\Cref{eq:parametric_y}); this completes the NDM generative process (\Cref{fig:graphical_model}).

\subsection{NDM instances}
\label{sec:model_instances}

To apply an NDM, we need to specify the distribution $f$ that is used to generate $\boldsymbol{y}$ (\Cref{eq:parametric_y}), and the link function $g(\cdot)$ to map the combination of the local parameters to the appropriate support of the parameters for $f$. \Cref{tab:link_functions} outlines example link functions and the corresponding distributions which they support.  

\begin{table}[ht]
\small
\centering
\begin{tabular}{ccc}
\toprule
\textbf{Link function $g$} &  \textbf{$g(x)$} & \textbf{Distributions $f$} \\
\midrule
identity & x & Normal, log-Normal \\
soft-plus & $\log(1+e^x)$ & Poisson, Gamma\\
exponential & $e^x$ & Poisson, Gamma \\
sigmoid & $\frac{1}{1 + e^{-x}}$ & Beta \\
inverse exponential & $e^{-x}$ & Exponential \\
\bottomrule
\end{tabular}
\caption{Example link functions and the distributions that they support.}
\label{tab:link_functions}
\end{table}

The appropriate choice of $f$ depends on the nature of the observations.  For example, it makes sense to use a Poisson to model voting counts (\Cref{sec:dat_vote}) or discrete sports data.  If one turns those data into percentages, however, it might make more sense to specify $f$ as a beta distribution.  The gamma, log-normal, and exponential distributions are natural choices for positive real-valued data.  In all cases, the exact choice should be made based on domain knowledge of the underlying processes.

\section{Inference}
\label{sec:inference}
In this section, we parallel the structure of \Cref{sec:model_family} by beginning with an inference algorithm for the parametric variant of the deconvolution model family (\Cref{sec:parametric_inference}).  We then design split and merge operations to construct the inference algorithm for the full nonparametric deconvolution model family (\Cref{sec:ndm_inference}).  We have released open source software for our model and inference methods at \url{https://github.com/ajbc/ndm}.

\subsection{Inference for Parametric Deconvolution Models}
\label{sec:parametric_inference}
Our central computational problem is inference: given the observed data $\boldsymbol{y}$, how do we determine the best values for the latent parameters in our model? In particular, inference involves estimating the latent variables and parameters---the global proportions $\boldsymbol{\beta}$, local proportions $\boldsymbol{\pi}$, global feature means $\boldsymbol{\mu}$ and covariances $\boldsymbol{\Sigma}$, local features $\boldsymbol{\bar{x}}$, and local counts $\boldsymbol{P}$.  As the true posterior for our model is intractable to compute, we approach this problem with variational inference~\citep{blei2017variational, Wainwright:2008}.

Variational inference finds a candidate approximation $q$ from a family of densities $\mathcal{Q}$ that is close the true posterior distribution $p$ by finding the $q$ that minimizes the KL divergence from the approximation $q\in\mathcal{Q}$ to the posterior $p$.  Using standard convexity arguments, this is equivalent to maximizing the evidence lower bound (ELBO), which is also called the variational objective:
\begin{equation}
    \mathcal{L}(q)  = \E_{q(\boldsymbol{\beta}, \boldsymbol{\pi}, \boldsymbol{\mu}, \boldsymbol{\Sigma}, \boldsymbol{\bar{x}}, \boldsymbol{P})}\left[\log p(\boldsymbol{y}, \boldsymbol{\beta}, \boldsymbol{\pi}, \boldsymbol{\mu}, \boldsymbol{\Sigma}, \boldsymbol{\bar{x}}, \boldsymbol{P}) - \log q(\boldsymbol{\beta}, \boldsymbol{\pi}, \boldsymbol{\mu}, \boldsymbol{\Sigma}, \boldsymbol{\bar{x}}, \boldsymbol{P})\right].
    \label{eq:ELBO}
\end{equation}
Here we define the family of approximating distributions $q\in\mathcal{Q}$ using the mean field assumption:
\begin{equation}
    q\left(\boldsymbol{\beta}, \boldsymbol{\pi}, \boldsymbol{\mu}, \boldsymbol{\Sigma}, \boldsymbol{\bar{x}}, \boldsymbol{P}\right) = 
    q\left(\boldsymbol{\beta}\right) ~\prod_{n=1}^N \Big[ q\left(\boldsymbol{\pi}_{n}\right) q\left(P_{n}\right) \Big]
    \prod_{k=1}^K \left[ q\left(\boldsymbol{\mu}_k\right)~q\left(\boldsymbol{\Sigma}_k\right) ~\prod_{n=1}^N q\left(\boldsymbol{\bar{x}}_{n,k}\right)\right],
\end{equation}
where each variable-specific approximation $q$ is parameterized by free variational parameters $\boldsymbol{\lambda}$.
For the factor proportions, the global proportion approximation $q(\boldsymbol{\beta})$ is a Dirichlet distribution with a $K$-dimensional free variational parameter vector $\boldsymbol{\lambda}[\beta]$; local proportion approximations $q(\boldsymbol{\pi}_n)$ are also Dirichlet distributions, each with a $K$-dimensional variational parameter vector $\boldsymbol{\lambda}[\pi_{n}]$.
For the factor descriptions, the global factor feature mean approximations $q(\boldsymbol{\mu_{k}})$ are Gaussian distributions with variational parameters $\boldsymbol{\lambda}[\mu_{k}]$; global factor feature covariance approximations $q(\boldsymbol{\Sigma}_{k})$ are inverse Wishart distributions with variational parameters $\boldsymbol{\lambda}[\Sigma_{k}]$. Local factor feature approximations $q(\boldsymbol{\bar{x}}_{n,k})$ are Gaussian distributions with variational parameters $\boldsymbol{\lambda}[{\bar{x}}_{n,k}]$, and
local count approximations $q(P_{n})$ are Poisson distributions, each with variational parameter $\lambda[{P}_{n}]$.

To maximize the ELBO (\Cref{eq:ELBO}), we need to be able to compute the expectations of the hidden parameters under $q$. The expectations for global factor feature means $\boldsymbol{\mu}$ and covariances $\boldsymbol{\Sigma}$ have analytic forms, but the remaining parameters ($\boldsymbol{\beta}, \boldsymbol{\pi}, \boldsymbol{\bar{x}}, \boldsymbol{P}$) do not. In lieu of analytic estimates for these second set of parameters, we use ``black box'' variational inference techniques \citep{Ranganath:2015}.  We construct our inference algorithm (\Cref{alg:parametric_inference}) by iterating over each of the parameters and latent variables ($\boldsymbol{\beta}, \boldsymbol{\pi}, \boldsymbol{\mu}, \boldsymbol{\Sigma}, \boldsymbol{\bar{x}}, \boldsymbol{P}$) and updating the corresponding variational parameters $\boldsymbol{\lambda}$ according to either analytic or black box estimates; we continue iterating over each parameter until convergence, giving us a coordinate ascent variational inference approach.  The remainder of this section will detail the estimation techniques for each parameter and outline the resulting algorithm.

\paragraph{Updates for global factor feature means $\boldsymbol{\mu}$ and covariances $\boldsymbol{\Sigma}$.}

To compute the expectations of the global factor feature means $\boldsymbol{\mu}_k$ for each factor $k$, we will derive the complete conditional distribution of these parameters, or $p(\boldsymbol{\mu}_k \g \boldsymbol{y}, \boldsymbol{\beta}, \boldsymbol{\pi}, \boldsymbol{\Sigma}, \boldsymbol{\bar{x}}, \boldsymbol{P})$.
These updates are straightforward given the conjugate relationships between the relevant distributions, and we obtain the complete conditional distribution
\begin{multline}
    p(\boldsymbol{\mu}_k \g \boldsymbol{y}, \boldsymbol{\beta}, \boldsymbol{\pi}, \boldsymbol{\Sigma}, \boldsymbol{\bar{x}}, \boldsymbol{P})
    = \\ \mathcal{N}_J\left( \left(\sigma_0^{-1} \mathbf{I}_M + \sum_{n=1}^N P_n \pi_{n,k} \boldsymbol{\Sigma}_k^{-1} \right)^{-1} \left(\sigma_0^{-1} \mathbf{I}_M \mu_0 + \boldsymbol{\Sigma}_k^{-1} \sum_{i=1}^N P_n \pi_{n,k} \boldsymbol{\bar{x}}_{n,k} \right), \right.\\
     \left. ~\left(\sigma_0^{-1} \mathbf{I}_M + \sum_{i=1}^N  P_n \pi_{n,k}\boldsymbol{\Sigma}_k^{-1} \right)^{-1}\right).
    \label{eq:complete_conditional_mu_assumption}
\end{multline}
Similarly, we obtain the complete conditional distributions for global covariances $\boldsymbol{\Sigma}_k$:
\begin{multline}
    p(\boldsymbol{\Sigma}_k \g \boldsymbol{y}, \boldsymbol{\beta}, \boldsymbol{\pi}, \boldsymbol{\mu}, \boldsymbol{\bar{x}}, \boldsymbol{P})
    = \\
    \mathcal{W}^{-1}\left(\nu_0 + \sum_{n=1}^N P_n \pi_{n,k}, 
    ~\boldsymbol{\Psi}_0 + \sum_{n=1}^N P_n \pi_{n,k} \left(\boldsymbol{\bar{x}}_{n,k}-\boldsymbol{\mu}_{k}\right)\left(\boldsymbol{\bar{x}}_{n,k}-\boldsymbol{\mu}_{k}\right)^\top\right).
    \label{eq:complete_conditional_Sigma_assumption}
\end{multline}
To update the estimates of $\boldsymbol{\mu}_k$ and $\boldsymbol{\Sigma}_k$, we set the variational parameters $\boldsymbol{\lambda}[\mu_k]$ and $\boldsymbol{\lambda}[\Sigma_k]$ to be the values of their corresponding terms in these complete condition distributions, using the current expectations of all of the parameters in the conditioning set.\footnote{While we can infer the scale parameter $\sigma$ for global factor features $\boldsymbol{\mu}$, we opt to fix this at a low value and only infer the means; thus our estimates for $\boldsymbol{\mu}$ are nearly point estimates.  This speeds up convergence and gives us better estimates of local factor feature variances, which are more important in using the fitted result to answer questions about real-world data.}   For example, the degrees of freedom parameter $\nu$ for covariance matrix $\boldsymbol{\Sigma}_k$ is estimated to be
\begin{equation}
    \boldsymbol{\lambda}[\Sigma_k(\nu)] =  \nu_0 + \sum_{n=1}^N \E_q[P_n]~ \E_q[\pi_{n,k}].
\end{equation}

\paragraph{Black box variational inference overview.}  To estimate the remaining parameters, we turn to black box variational inference techniques \citep{Ranganath:2015}.  We will describe this approach using generic latent variable $z$ with variational parameter $\lambda[z]$.

Recall that our objective is to maximize the ELBO (\Cref{eq:ELBO}); black box variational inferences relies on stochastic optimization to do this.  We want to approximate the true gradient of the ELBO, which we express as an expectation with respect to the variational distribution.  The true gradient with respect to a generic variational parameter $\lambda[z]$ is written as 
\begin{equation}
    \nabla_{\lambda[z]} \mathcal{L} =
    \E_q \left[ \nabla_{\lambda[z]} \log q(z \g \boldsymbol{\lambda}[z]) \left( \log p^{z}(\boldsymbol{y}, z, \dots) - \log q(z \g \boldsymbol{\lambda}[z])\right)\right].
\end{equation}
This gradient expressions contain a term for the log probability of all terms containing the hidden parameter of interest, or $\log p^{z}$.  For example, the log probability for local features $\boldsymbol{\bar{x}}_{n,k}$ is defined as follows:
\begin{equation}
    \log p^{\bar{x}}_{n,k}( \boldsymbol{y}, \boldsymbol{\Sigma}, \boldsymbol{\beta}, \boldsymbol{\pi}, \boldsymbol{\mu}, \boldsymbol{\bar{x}}, \boldsymbol{P}) \triangleq \log p(\boldsymbol{\bar{x}}_{n,k} \g \boldsymbol{\mu}_k, \boldsymbol{\Sigma}_k, \boldsymbol{\pi}_n, P_n) + \log p(y_{n} \g \boldsymbol{\bar{x}}_{n}, \boldsymbol{\pi}_n).
    \label{eq:partiallog}
\end{equation}

The objective is now to approximate the gradient $\nabla_{\lambda[z]} \mathcal{L}$ using $S$ samples from the variational distribution: $z[s] \sim q(z \g \lambda[z])$.  Using these samples, we construct the following noisy unbiased estimate of the gradient with respect to variational parameter $\lambda[z]$:
\begin{equation}
    \nabla_{\lambda[z]} \mathcal{L} \approx
    \tilde\nabla_{\lambda[z]} \mathcal{L} = 
    \frac{1}{S} \sum_{s=1}^S \left[ \nabla_{\lambda[z]} \log q(z[s] \g \lambda[z]) \left( \log p^z(\boldsymbol{y}, z[s], \dots) - \log q(z[s] \g \lambda[z])\right) \right].
\end{equation}

Once we have a noisy estimate of the gradient, we can update the corresponding variational parameter in the standard stochastic gradient ascent manner; at iteration $t$ this update is
\begin{equation}
    \lambda_{t+1} = \lambda_t + \rho_t \tilde\nabla_{\lambda} \mathcal{L},
    \label{eq:lambdaupdate}
\end{equation}
where the learning rate $\rho_t$ meets the Robbins-Monro conditions,
\begin{equation}
    \sum_{t=1}^\infty \rho_t = \infty \qquad
    \mbox{and} \qquad
    \sum_{t=1}^\infty (\rho_t)^2 < \infty.
\end{equation}

As noted by \citet{Ranganath:2015}, the variance of the black box estimator of the gradient can be large; this poses a challenge to achieving convergence in a reasonable time frame.  To control the variance of the gradient, we use approaches suggested by \citet{Ranganath:2015}, including control variates and RMSProp \citep{tieleman2012lecture} (in lieu of AdaGrad \citep{duchi2011adaptive}).

\paragraph{Black box updates for the remaining parameters ($\boldsymbol{\beta}$, $\boldsymbol{\pi}$, $\boldsymbol{\bar{x}}$,  $\boldsymbol{P}$).}
Following the black box variational inference framework, updates for the remaining parameters are straightforward.  \Cref{app:partiallogjoints} lists the log probabilities containing only the parameters of interest (e.g., \Cref{eq:partiallog}),  \Cref{app:gradients} contains the gradients of all log $q$ distributions used in inference, and details on how we set learning rates can be found in \Cref{app:learningrates}.
Readers who are interested in seeing additional details, beyond what is supplied in the appendix, are invited to explore our open-source implementation of the algorithm (\url{https://github.com/ajbc/ndm}).

To generalize inference for a wide range of variants in the deconvolution model family, we only need to update the log probability terms (\Cref{app:partiallogjoints}; e.g., \Cref{eq:partiallog}) that contain $p(\boldsymbol{y}_{n} \g \boldsymbol{\bar{x}}_{n}., \boldsymbol{\pi}_n)$. or $\log p^{\bar{x}}$ and $\log p^{\pi}$.  Here, we simply update the $p(\boldsymbol{y}_{n})$ likelihood term with the distribution $f$ and link function $g$ for the given model instance, and no other changes are needed for inference.

Both these black box updates and the analytic updates for global factor means $\boldsymbol{\mu}$ and covariances $\boldsymbol{\Sigma}$ are combined to give us the full parametric inference algorithm (\Cref{alg:parametric_inference}).

\LinesNumbered
\begin{algorithm}
\small
\DontPrintSemicolon
\SetNoFillComment
\SetAlgoNoEnd
\KwIn{observations $y$}
\KwOut{approximate posterior $q$ parameterized by $\boldsymbol{\lambda}$}
\textbf{Initialize} variational parameters $\boldsymbol{\lambda}$ (\Cref{app:initialize}) \;
\textbf{Initialize} iteration count $t = 0$ \;
\tcc{See \Cref{app:partiallogjoints,app:gradients} for definitions of $p^z$ and $\nabla_\lambda \log q$.}
\While {change in ELBO $< \delta$ 
    {\normalfont (\Cref{app:convergence})}}{
        set learning rates $\boldsymbol{\rho}_t$ (\Cref{app:learningrates}) \;
        
        \BlankLine
        \For{observation $n=1:N$}{
            \tcc{update local factors $\boldsymbol{\bar{x}}_n$}   
            \For{factor $k=1:K$}{
                \For{sample $s=1:S$}{
                    sample $\boldsymbol{\bar{x}}_{n,k}[s] \sim q(\boldsymbol{\bar{x}}_{n,k} \g \boldsymbol{\lambda}[\bar{x}_{n,k}])$ \tcc*[l]{$q$ is $M$ univariate Gaussians}
                }
                $\boldsymbol{\lambda}[\bar{x}_{n,k}(\mu)] \pluseq \rho^{\bar{x}(\mu)}_t
                    \frac{1}{S}\sum_{s=1}^S \left[ \nabla_{\boldsymbol{\lambda}[\bar{x}_{n,k}(\mu)]} \log q(\boldsymbol{\bar{x}}_{n,k}[s] \g \boldsymbol{\lambda}) \left( \log p^{\bar{x}}(\boldsymbol{y}, \boldsymbol{\bar{x}}_{n,k}[s], \dots) - \log q(\boldsymbol{\bar{x}}_{n,k}[s] \g \boldsymbol{\lambda}\right) \right] $ \;
                $\boldsymbol{\lambda}[\bar{x}_{n,k}(\sigma)] \pluseq \rho^{\bar{x}(\sigma)}_t
                    \frac{1}{S}\sum_{s=1}^S \left[ \nabla_{\boldsymbol{\lambda}[\bar{x}_{n,k}(\sigma)]} \log q(\boldsymbol{\bar{x}}_{n,k}[s] \g \boldsymbol{\lambda}) \left( \log p^{\bar{x}}(\boldsymbol{y}, \boldsymbol{\bar{x}}_{n,k}[s], \dots) - \log q(\boldsymbol{\bar{x}}_{n,k}[s] \g \boldsymbol{\lambda}\right) \right] $ \;
            }
            
            \BlankLine
            \tcc{update local proportions $\boldsymbol{\pi}_n$}
            \For{sample $s=1:S$}{
                sample $\boldsymbol{\pi}_{n}[s] \sim q(\boldsymbol{\pi}_{n} \g \boldsymbol{\lambda}[\pi_{n}])$ \tcc*[l]{$q$ is a $K$-dimensional Dirichlet}
            }
            $\boldsymbol{\lambda}[\pi_{n}] \pluseq \rho^{\pi}_t
                    \frac{1}{S}\sum_{s=1}^S \left[ \nabla_{\boldsymbol{\lambda}[\pi_{n}]} \log q(\boldsymbol{\pi}_{n}[s] \g \boldsymbol{\lambda}) \left( \log p^{\pi}(\boldsymbol{y}, \boldsymbol{\pi}_{n}[s], \dots) - \log q(\boldsymbol{\pi}_{n}[s] \g \boldsymbol{\lambda}\right) \right] $ \;
            
            \BlankLine
            \tcc{update local counts $P_n$}
            \For{sample $s=1:S$}{
                sample $P_{n}[s] \sim q(P_{n} \g \lambda[P_{n}])$ \tcc*[l]{$q$ is Poisson}
            }
            $\lambda[P_n] \pluseq \rho^{P}_t
                    \frac{1}{S}\sum_{s=1}^S \left[ \nabla_{\boldsymbol{\lambda}[P_{n}]} \log q(P_{n}[s] \g \boldsymbol{\lambda}) \left( \log p^{P}(\boldsymbol{y}, P_{n}[s], \dots) - \log q(P_{n}[s] \g \boldsymbol{\lambda}\right) \right] $ \;
        }
        
        \BlankLine
        \tcc{update global factor feature means $\boldsymbol{\mu}$ and covariances $\boldsymbol{\Sigma}$}
        \For{factor $k=1:K$}{
            $\boldsymbol{\lambda}[\mu_k(\mu)] = \left(\sigma_0^{-1} \mathbf{I}_M + \sum_{n=1}^N \E_q[P_n] ~\E_q[\pi_{n,k}] ~\E_q[\boldsymbol{\Sigma}_k]^{-1} \right)^{-1} \left(\sigma_0^{-1} \mathbf{I}_M \mu_0 + \E_q[\boldsymbol{\Sigma}_k]^{-1} \sum_{n=1}^N \E_q[P_n] ~\E_q[\pi_{n,k}]~\E_q[\boldsymbol{\bar{x}}_{n,k}] \right)$ \;

            $\boldsymbol{\lambda}[\Sigma_k(\nu)] = \nu_0 + \sum_{n=1}^N \E_q[P_n]~ \E_q[\pi_{n,k}]$ \;
            
            $\boldsymbol{\lambda}[\Sigma_k(\Psi)] = \boldsymbol{\Psi}_0 + \sum_{n=1}^N \E_q[P_n]~ \E_q[\pi_{n,k}] \left(\E_q[\boldsymbol{\bar{x}}_{n,k}]-\E_q[\boldsymbol{\mu}_{k}]\right)\left(\E_q[\boldsymbol{\bar{x}}_{n,k}]-\E_q[\boldsymbol{\mu}_{k}]\right)^\top$ \;
        }
        
        \BlankLine
        \tcc{update global proportions $\boldsymbol{\beta}$}
        \For{sample $s=1:S$}{
            sample $\boldsymbol{\beta}[s] \sim q(\boldsymbol{\beta} \g \boldsymbol{\lambda}[\beta])$ \tcc*[l]{$q$ is a $K$-dimensional Dirichlet}
        }
        $\boldsymbol{\lambda}[\beta] \pluseq \rho^{\beta}_t
                \frac{1}{S}\sum_{s=1}^S \left[ \nabla_{\boldsymbol{\lambda}[\beta]} \log q(\boldsymbol{\beta}[s] \g \boldsymbol{\lambda}) \left( \log p^{\beta}(\boldsymbol{y}, \boldsymbol{\beta}[s], \dots) - \log q(\boldsymbol{\beta}[s] \g \boldsymbol{\lambda}\right) \right] $ \;
        
    \BlankLine
    update iteration count $t \pluseq 1$ \;
}
\BlankLine
\Return{$\boldsymbol{\lambda}$} \;
\caption{Variational inference algorithm for parametric deconvolution models}
\label{alg:parametric_inference}
\end{algorithm}

\subsection{Inference for NDMs}
\label{sec:ndm_inference}

Now that we have established an efficient inference algorithm for the parametric version of deconvolution models, we turn to split and merge procedures to assist us with estimating the latent variables in the nonparametric context. While many split-merge procedures exist \citep{ueda1999smem, jain2004split, dahl2005sequentially, wang2012split}, \citet{bryant2012truly} introduced split and merge procedures for inference in the Hierarchical Dirichlet Process with an online variational inference algorithm; we adapt these procedures for our model.

The core idea of this approach is to treat the parametric algorithm (\Cref{alg:parametric_inference}) as a batch with a fixed number of factors $K$, with one major exceptions: instead of being a $K$-dimensional vector, the global proportions $\boldsymbol{\beta}$ are instead a $(K+1)$-dimensional vector; the last element accounts for the mass of all remaining factors $k>K$.  Once inference has converged at the batch level, factors can be split (creating new ones) or merged (merging redundant pairs).  Then, further batches can be run until the number of factors and the associated parameters for those factors converge.

\paragraph{Split operation (creating new factors).}  The split operation allows splitting a factor $k$ into two factors, $k'$ and $k''$.  To summarize: given the current variational approximation $q$ and corresponding variational parameters $\boldsymbol{\lambda}$, we first initialize the variational parameters $\boldsymbol{\lambda}^S$ for the candidate approximation $q^S$, taking care to introduce small amounts of random noise so that the new factors can distinguish themselves from each other.
Then, we run a single iteration of the batch algorithm (\Cref{alg:parametric_inference}) to update the new candidate variational parameters $\boldsymbol{\lambda}^S$.  Last, we accept the split candidate approximation $q^S$ if it increases the ELBO (\Cref{eq:ELBO}), and reject it otherwise.

We initialize the candidate variational parameters for new factors $k'$ and $k''$ as follows. Both global and local proportions ($\boldsymbol{\beta}$ and $\boldsymbol{\pi}$, respectively) are split between the two new factors, using a rate $\rho^{\textsc{sm}}_t$ to determine how the proportions are divided between the two new factors. We set $\rho^{\textsc{sm}}_t= (t + 4) ^ {-0.5}$ for iteration $t$.  For the global proportions $\boldsymbol{\beta}$, the variational parameters are initialized to
\begin{equation}
     \boldsymbol{\lambda}^S[{\beta}_{k'}] = \rho^{\textsc{sm}}_t  \boldsymbol{\lambda}[{\beta}_{k}]
        \qquad \mbox{and} \qquad
    \boldsymbol{\lambda}^S[{\beta}_{k''}] = (1 - \rho^{\textsc{sm}}_t) \boldsymbol{\lambda}[{\beta}_{k}].
    \label{eq:S1}
\end{equation}
For local proportions $\boldsymbol{\pi}_n$, for all observations $n=1,\dots,N$, the variational parameters are initialized to
\begin{equation}
    \boldsymbol{\lambda}^S[\pi_{n,k'}] = \rho^{\textsc{sm}}_t  \boldsymbol{\lambda}[\pi_{n,k}]
        \qquad \mbox{and} \qquad
    \boldsymbol{\lambda}^S[\pi_{n,k''}] = (1 - \rho^{\textsc{sm}}_t) \boldsymbol{\lambda}[{\pi}_{n,k}].
    \label{eq:S2}
\end{equation}

To break the symmetry between the two new factors, we must introduce a small amount of noise for the global factor features $\boldsymbol{\mu}$; thus we can initialize the variational parameters for the factor features to
\begin{equation}
    \boldsymbol{\lambda}^S[\mu_{k'}] = \boldsymbol{\lambda}[\mu_{k}]
        \qquad \mbox{and} \qquad
   \boldsymbol{\lambda}^S[\mu_{k''}] = \boldsymbol{\lambda}[\mu_{k}] + \boldsymbol{\varepsilon},
   \label{eq:S3}
\end{equation}
where $\boldsymbol{\varepsilon}$ is an $M$-dimensional vector where each element is drawn from a Gaussian distribution, or $\varepsilon_m\sim\mathcal{N}(0,\sigma)$ with small scale $\sigma$.  Alternatively, we can run a simple clustering algorithm (e.g., K-means) with two clusters and treat the expectations of the local factor feature means for factor $k$, or $\E[\boldsymbol{\lambda}[\bar{x}_k(\mu)]$, as input ``data;'' this approach performs well in practice and does not require defining a scale $\sigma$ hyper-parameter, to which the split operation would be sensitive; thus we use a K-means approach.
  
Given that the symmetry between the factors is broken with global factor features $\boldsymbol{\mu}$, we can simply carry over the variational parameters for the global factor covariances $\boldsymbol{\Sigma}$, giving us straightforward initializations,
\begin{equation}
    \boldsymbol{\lambda}^S[\Sigma_{k'}(\nu)] = \boldsymbol{\lambda}^S[\Sigma_{k''}(\nu)] =
    \boldsymbol{\lambda}[\Sigma_{k}(\nu)]
        \qquad \mbox{and} \qquad
    \boldsymbol{\lambda}^S[\Sigma_{k'}(\Psi)] = \boldsymbol{\lambda}^S[\Sigma_{k''}(\Psi)] =
    \boldsymbol{\lambda}[\Sigma_{k}(\Psi)].
    \label{eq:S4}
\end{equation}

Local factor features $\boldsymbol{\bar{x}}$ can similarly be copied; for all observations $n=1,\dots,N$, the variational parameters are initialized as
\begin{equation}
    \boldsymbol{\lambda}^S[\bar{x}_{n,k'}(\mu)] = 
    \boldsymbol{\lambda}^S[\bar{x}_{n,k''}(\mu)] =  \boldsymbol{\lambda}[\bar{x}_{n,k}(\mu)]
    \quad \mbox{and} \quad
    \boldsymbol{\lambda}^S[\bar{x}_{n,k'}(\sigma)] = 
    \boldsymbol{\lambda}^S[\bar{x}_{n,k''}(\sigma)] =  \boldsymbol{\lambda}[\bar{x}_{n,k}(\sigma)].
    \label{eq:S5}
\end{equation}

No other variational parameters are impacted by the split operation during initialization; all remaining $\boldsymbol{\lambda}^S$ for the candidate $q^S$ are initialized by copying over their values from the current variational parameters $\boldsymbol{\lambda}$.
Once all the variational parameters have been initialized, we run a single iteration, or ``trial iteration,'' of the batch algorithm (\Cref{alg:parametric_inference}, lines 4--23\footnote{\label{note:k_plus_1} Global proportions $\boldsymbol{\beta}$ need to be modified to be $(K+1)$-dimensional, which impact lines 13, and 21--23.  Line 13 need only use the first $K$ elements of $\boldsymbol{\beta}$ in computing $\log p^\pi$ and is otherwise the same.  Lines 21--23 are impacted by updating $\boldsymbol{\lambda}[\beta]$ to be $(K+1)$-dimensional; then, sampling on line 22 and using the samples on line 23 are both straightforward. }, returning $\boldsymbol{\lambda}^S$), which updates each variational parameter $\boldsymbol{\lambda}^S$ exactly once.  Now, we can compute the ELBO ($\mathcal{L}$, \Cref{eq:ELBO}) of the candidate approximation $q^S$ and compare it to $q$; if the ELBO of the split candidate $q^S$ is larger than that of the current approximation $q$, or $\mathcal{L}(q^S) > \mathcal{L}(q)$, then we accept the split candidate approximation $q^S$ and continue the inference algorithm with an additional factor. 

When the splitting stage is triggered, all $K$ factors that exist at the start of the stage are considered for splitting (ordered randomly).  Each of the $K$ factors is considered individually during this stage: each factor $k$ goes through the split operation as just described.  When the splitting stage has completed, the current approximation $q$ can have at most $2K$ factors.

\paragraph{Merge operation (removing redundant factors).}
The merge operation considers two factors $k'$ and $k''$ to combine into a single factor $k$.  This procedure is similar to the split operation: first we initialize the variational parameters $\boldsymbol{\lambda}^M$ for the candidate approximation $q^M$, then we update the variational parameters $\boldsymbol{\lambda}^M$ with a single iteration of the batch algorithm, and accept or reject the merge candidate approximation $q^M$ based on the ELBO.

We initialize the candidate variational parameters for the new factor $k$ as follows.  Both global and local proportions ($\boldsymbol{\beta}$ and $\boldsymbol{\pi}$, respectively) are summed,
\begin{equation}
     \boldsymbol{\lambda}^M[{\beta}_{k}] =  \boldsymbol{\lambda}[{\beta}_{k'}] + \boldsymbol{\lambda}[{\beta}_{k''}]
        \qquad \mbox{and} \qquad
    \boldsymbol{\lambda}^M[{\pi}_{n,k}] = 
    \boldsymbol{\lambda}[{\pi}_{n,k'}] +  
    \boldsymbol{\lambda}[{\pi}_{n,k''}],
    \label{eq:M1}
\end{equation}
for all observations $n=1,\dots,N$.  The other global variational parameters are initialized based on weighted averages of the two original factors $k'$ and $k''$ (the proportions $\boldsymbol{\beta}$ or $\boldsymbol{\pi}$ being the weights).  For global factor feature distribution parameters $\boldsymbol{\mu}$, we have
\begin{equation}
     \boldsymbol{\lambda}^M[\mu_{k}] =  \frac{\boldsymbol{\lambda}[\beta_{k'}]\boldsymbol{\lambda}[\mu_{k'}] + \boldsymbol{\lambda}[\beta_{k''}]\boldsymbol{\lambda}[\mu_{k''}]}{
     \boldsymbol{\lambda}[\beta_{k'}] + \boldsymbol{\lambda}[\beta_{k''}]}.
     \label{eq:M2}
\end{equation}

For global factor feature distribution covariances $\boldsymbol{\Sigma}$, the variational parameters are initialized as
\begin{equation}
     \boldsymbol{\lambda}^M[\Sigma_{k}(\nu)] =  \frac{\boldsymbol{\lambda}[\beta_{k'}]\boldsymbol{\lambda}[\Sigma_{k'}(\nu)] + \boldsymbol{\lambda}[\beta_{k''}]\boldsymbol{\lambda}[\Sigma_{k''}(\nu)]}{
     \boldsymbol{\lambda}[\beta_{k'}] + \boldsymbol{\lambda}[\beta_{k''}]}
     \label{eq:M3}
\end{equation}
and
\begin{equation}
     \boldsymbol{\lambda}^M[\Sigma_{k}(\Psi)] =  \frac{\boldsymbol{\lambda}[\beta_{k'}]\boldsymbol{\lambda}[\Sigma_{k'}(\Psi)] + \boldsymbol{\lambda}[\beta_{k''}]\boldsymbol{\lambda}[\Sigma_{k''}(\Psi)]}{
     \boldsymbol{\lambda}[\beta_{k'}] + \boldsymbol{\lambda}[\beta_{k''}]}.
     \label{eq:M4}
\end{equation}
Local factor feature values $\boldsymbol{\bar{x}}$ are initialized based on the weighted average of the two original factors; for all observations $n=1,\dots,N$, we set the variational parameters to
\begin{equation}
    \boldsymbol{\lambda}^M[\bar{x}_{n,k}(\mu)] =  \frac{\boldsymbol{\lambda}[\pi_{n,k'}]\boldsymbol{\lambda}[\bar{x}_{n,k'}(\mu)] + \boldsymbol{\lambda}[\pi_{n,k''}]\boldsymbol{\lambda}[\bar{x}_{n,k''}(\mu)]}{
    \boldsymbol{\lambda}[\pi_{n,k'}] + \boldsymbol{\lambda}[\pi_{n,k''}]}
    \label{eq:M5}
\end{equation}
and
\begin{equation}
    \boldsymbol{\lambda}^M[\bar{x}_{n,k}(\sigma)] =  \frac{\boldsymbol{\lambda}[\pi_{n,k'}]\boldsymbol{\lambda}[\bar{x}_{n,k'}(\sigma)] + \boldsymbol{\lambda}[\pi_{n,k''}]\boldsymbol{\lambda}[\bar{x}_{n,k''}(\sigma)]}{
    \boldsymbol{\lambda}[\pi_{n,k'}] + \boldsymbol{\lambda}[\pi_{n,k''}]}.
    \label{eq:M6}
\end{equation}

No other variational parameters are impacted by the merge operation during initialization; all remaining $\boldsymbol{\lambda}^M$ for the candidate $q^M$ are initialized by copying over their values from the current variational parameters $\boldsymbol{\lambda}$.
After initializing the variational parameters, we run a single iteration of the batch algorithm (\Cref{alg:parametric_inference}, lines 4--23\textsuperscript{\ref{note:k_plus_1}}, returning $\boldsymbol{\lambda}^M$) and compute the ELBO (\Cref{eq:ELBO}) of the candidate approximation $q^M$, or $\mathcal{L}(q^M)$, for comparison with the ELBO of the current $q$, or $\mathcal{L}(q)$.  If $\mathcal{L}(q^M) > \mathcal{L}(q)$, we accept the merge candidate approximation, setting $q=q^M$, and continue inference with $K-1$ factors ($K$ for $\boldsymbol{\beta}$).

When the merging operation is triggered, only a subset of factor pairs are considered as merge candidates.  For every possible pair of factors $k'$ and $k''$, we compute the covariance of the local proportions $\boldsymbol{\pi}$ over all observations.  When this covariance is greater than zero, the pair is added to the merge candidate list.  Candidate pairs are ordered based on covariance, with the highest covariance pairs being considered for merging first.  We considered 
identifying candidate pairs based on Euclidean distance of the factor means, but proportion covariance worked better in practice; it additionally has the advantages of being faster to compute and having a natural threshold (greater than zero).

\paragraph{Full inference algorithm.}  We combine the split and merge operations with the variational inference updates to get the full inference algorithm for nonparametric deconvolution models, shown in \Cref{alg:nonparametric_inference}.

\begin{algorithm}[ht]
\small
\DontPrintSemicolon
\SetNoFillComment
\SetAlgoNoEnd
\KwIn{observations $y$}
\KwOut{approximate posterior $q$ parameterized by $\boldsymbol{\lambda}$}
\textbf{Initialize} variational parameters $\boldsymbol{\lambda}$ (\Cref{app:initialize}) \;
\textbf{Initialize} number of factors $K$ \;
\textbf{Initialize} iteration count $t = 0$ \;
\While {not converged 
    {\normalfont (\Cref{app:convergence})}}{
    \textbf{Run batch} according to \Cref{alg:parametric_inference}, lines 3--24\textsuperscript{\ref{note:k_plus_1}} (until convergence) \;
    \BlankLine
    
    \tcc{Merge}
    set \emph{merge candidates} to be all $(k',k'')$ where $\mbox{cov}(\boldsymbol{\pi}_{k}, \boldsymbol{\pi}_{k''}) > 0$, ordered by covariance\;
    \For{factor pairs $(k',k'')\in$ merge candidates}{
        \textbf{Initialize} $\boldsymbol{\lambda}^M$ according to \Cref{eq:M1,eq:M2,eq:M3,eq:M4,eq:M5,eq:M6} and $\boldsymbol{\lambda}^M = \boldsymbol{\lambda}$ for all remaining \;
        \textbf{Run trial iteration} according to \Cref{alg:parametric_inference}, lines 4--23\textsuperscript{\ref{note:k_plus_1}} using $\boldsymbol{\lambda}^M$ \;
        \If{$\mathcal{L}(q(\boldsymbol{\lambda}^M)) > \mathcal{L}(q(\boldsymbol{\lambda}))$}{
            $\boldsymbol{\lambda} = \boldsymbol{\lambda}^M$ \;
            $K \minuseq 1$ \;
        }
    }\BlankLine
    
    \tcc{Split}
    \For{factor $k=1:K$}{
        \textbf{Initialize} $\boldsymbol{\lambda}^S$ according to \Cref{eq:S1,eq:S2,eq:S3,eq:S4,eq:S5} and $\boldsymbol{\lambda}^S = \boldsymbol{\lambda}$ for all remaining \;
        \textbf{Run trial iteration} according to \Cref{alg:parametric_inference}, lines 4--23\textsuperscript{\ref{note:k_plus_1}} using $\boldsymbol{\lambda}^S$ \;
        \If{$\mathcal{L}(q(\boldsymbol{\lambda}^S)) > \mathcal{L}(q(\boldsymbol{\lambda}))$}{
            $\boldsymbol{\lambda} = \boldsymbol{\lambda}^S$ \;
            $K \pluseq 1$ \;
        }
    }
}
\BlankLine
\Return{$\boldsymbol{\lambda}$} \;
\caption{Variational inference algorithm for nonparametric deconvolution models.}
\label{alg:nonparametric_inference}
\end{algorithm}

\section{Empirical Results}
\label{sec:results}

We evaluated the performance of NDMs trained on simulated data (\Cref{sec:dat_sim}) and on voting data (\Cref{sec:dat_vote}).  We show that modeling local features leads to improved estimates of parameters and latent variables,
and that a fitted NDM captures between-group variability better than existing models.
We begin by showing improved latent variable estimates with simulated data (\Cref{sec:dat_sim}). Then, we turn to addressing variation in demographic voting patterns across voting precincts with data from the 2016 election in California (\Cref{sec:dat_vote}).

\subsection{Simulations}
\label{sec:dat_sim}

A main purposes of applying a deconvolution model to data is to recover information that has been lost during the convolutional (or aggregation) process. We often do not have ground truth observations for each of the components in the convolutional process for applications of interest. Thus, we rely on simulations to provide data where the particles that we wish to recover from the aggregated data are known in order to validate our model. We also compare results from our deconvolution model with results from related models on these simulated data.

\begin{figure}
\centering
\includegraphics[width=6in]{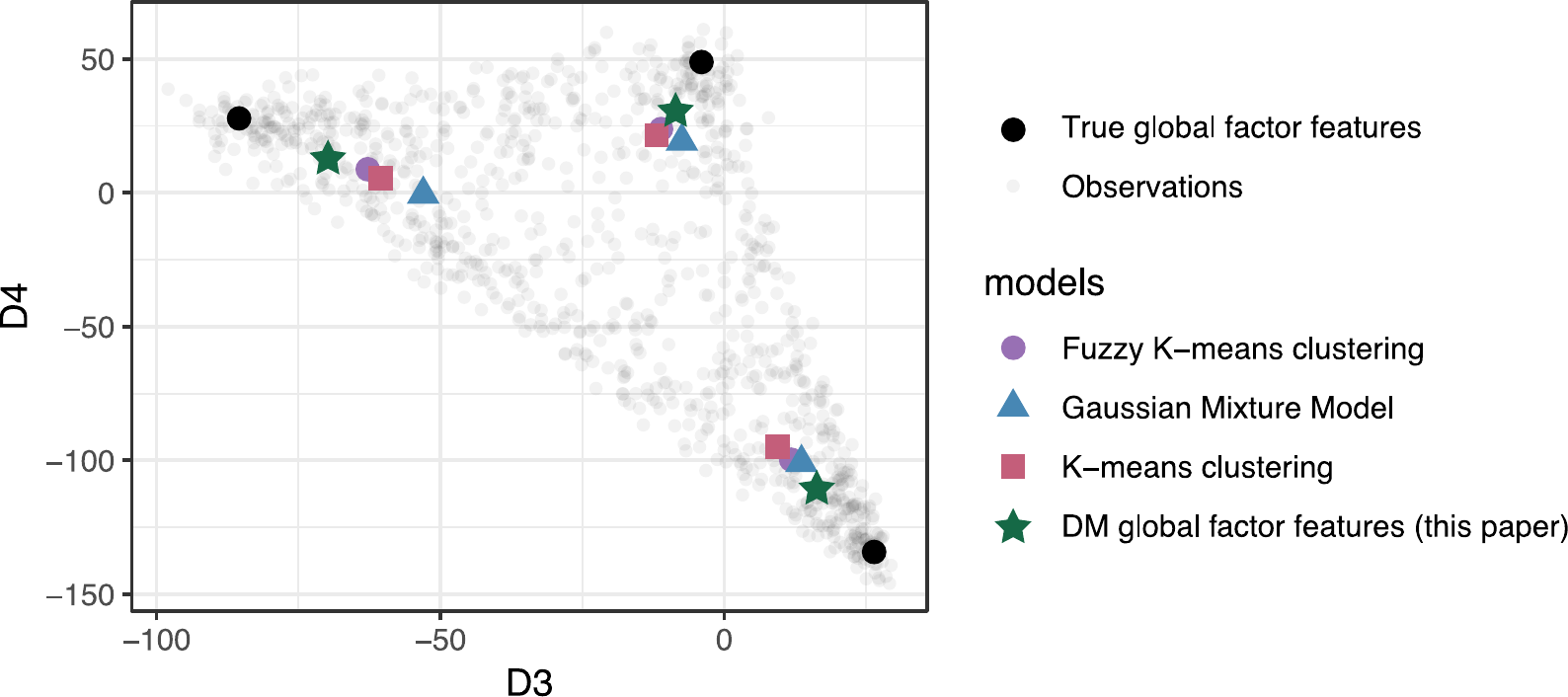}
\caption{The parametric deconvolution model (DM) discovers global factors closer to the true global factors in simulated data, as compared to standard clustering methods.  The simulated data contain 1,000 observations in five dimensions (results shown in dimensions 3 and 4, but are comparable in other dimensions).}
\label{fig:sim_3K}
\end{figure}

\paragraph{Simulated Data Description.}
We simulated data in four different ways in order to quantify performance for a variety of underlying data-generating processes; \Cref{sec:app_sim} provides more details on these simulation procedures.  Briefly, we generated data similar to the NDM generative process (\Cref{sec:model_family}), where individual particles are generated from observation-specific means (simulation procedure 1, \Cref{app:sim1}), or individual particles are generated directly from global means (simulation procedure 2, \Cref{app:sim2}).  We also modified the NDM generative process to add additional hierarchical complexity to the data by including some number of ``modes'' from which the particles are drawn; these modes are either associated with local factors (simulation procedure 3, \Cref{app:sim3}) or associated with each global factor (simulation procedure 4, \Cref{app:sim4}).

We simulated data from multiple distributions $f$ (and link functions $g$), including Gaussian (identity link, $g(x) = x$), Poisson (soft-plus link, $g(x) = \log(e^x + 1)$), beta (sigmoid link, $g(x) = 1\times10^{-6} + (1-2\times10^{-6}) / (1 + \exp(-10(x -0.5)))$; see \Cref{sec:app_sim}), and gamma (soft-plus link).
Except where stated otherwise, we simulated data with ten random seeds for each setting and report average performance across the ten seeds; we set the number of factors $K=10$, the number of observations $N=1000$, and the number of features $M=20$.

\paragraph{Comparison methods.}  We focus our comparisons on standard decomposition methods, as these are the most similar family of models to deconvolution models.  While we have introduced a nonparametric model family, we restrict our simulated evaluations to the parametric variant because parametric decomposition models are more readily available as comparison methods. In particular, we compare parametric deconvolution models (DM) to 
factor analysis~\citep[FA;][]{harman1960modern},
principal component analysis~\citep[PCA;][]{hotelling1933analysis},
non-negative matrix factorization~\citep[NMF;][]{Lee01},
Gamma-Poisson matrix factorization~\citep[GaP;][]{CannyGaP}, and
Gaussian probabilistic matrix factorization~\citep[PMF;][]{PMF}.
When available, we used the \texttt{scikit-learn} Python library decomposition module \cite{scikit-learn} with default parameter settings; PMF and GaP required additional implementations.\footnote{We relied on the \texttt{ProbabilisticMatrixFactorization} library for PMF (\url{https://github.com/fuhailin/Probabilistic-Matrix-Factorization}) and our own implementation of GaP.  A framework to run the comparison methods is included in the released software.}
Some of the simulated data sets are incompatible with certain comparison methods; for instance, GaP can only be applied to integer data.  In these cases, irrelevant comparison methods are omitted.

\begin{figure}[tbh]
\centering
\includegraphics[width=\textwidth]{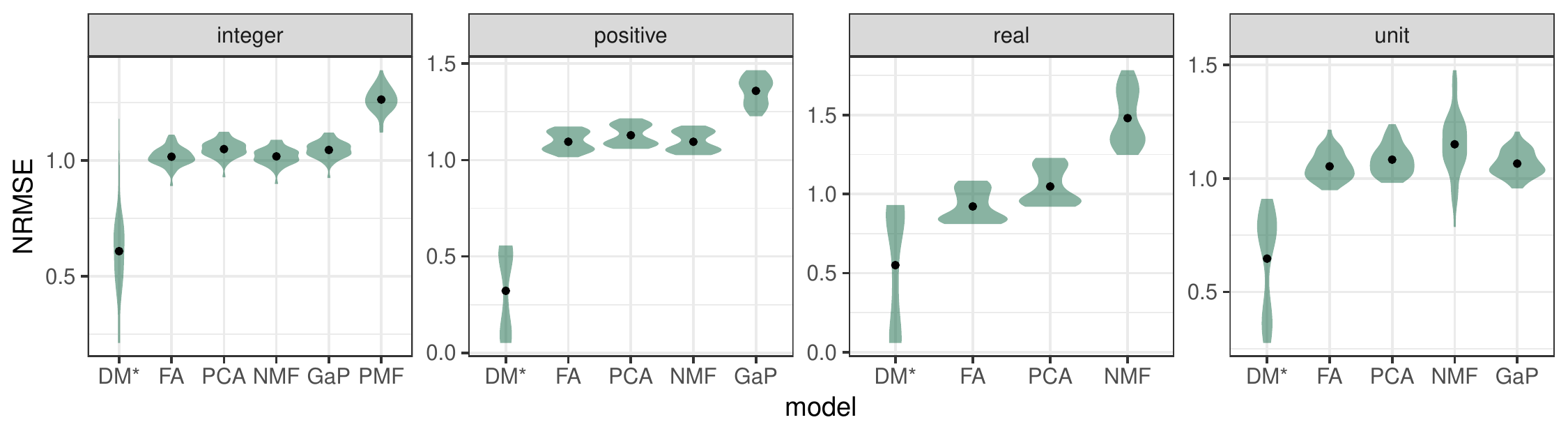}
\caption{Normalized root mean square error (NRMSE, lower is better) of estimated global factor feature means $\hat{\boldsymbol{\mu}}$ from the true simulated feature means $\boldsymbol{\mu}$ (\Cref{eq:NRMSE}), grouped by simulated data domain and averaged across all seeds and simulation procedures. Our model family, deconvolution models (DM; starred) outperforms all other methods in the comparison.
}
\label{fig:NRMSE_mu}
\end{figure}

\noindent\paragraph{Estimating global factor feature distributions and proportions.}
We compared estimates of global factor feature distributions across methods on our simulated data.  To do this, we fit DMs and our comparative models to the simulated data and compared point estimates of the global factor distributions to the generated values using normalized root mean square error (NRMSE; we normalize to allow for averaging across data sets) of the estimated global factor feature distributions $\hat{\boldsymbol{\mu}}$ from the true simulated features $\boldsymbol{\mu}$, or
\begin{equation}
\mbox{NRMSE}(\hat{\boldsymbol{\mu}}) = \frac{\sqrt{\frac{\sum_{k=1}^K\sum_{m=1}^M (\hat{\mu}_{k,m} - \mu_{k,m})^2 }{K\times M}}}{\max(\boldsymbol{\mu}) - \min(\boldsymbol{\mu})}.
\label{eq:NRMSE}
\end{equation}
We find that our approach both recovers good estimates of the local factors and also improves upon the global factor estimates over related methods (\Cref{fig:NRMSE_mu}). This suggests that augmenting the distributions of a deconvolutional model to include local distributions improves the estimates of the global distributions.

\begin{figure}
\centering
\includegraphics[width=\textwidth]{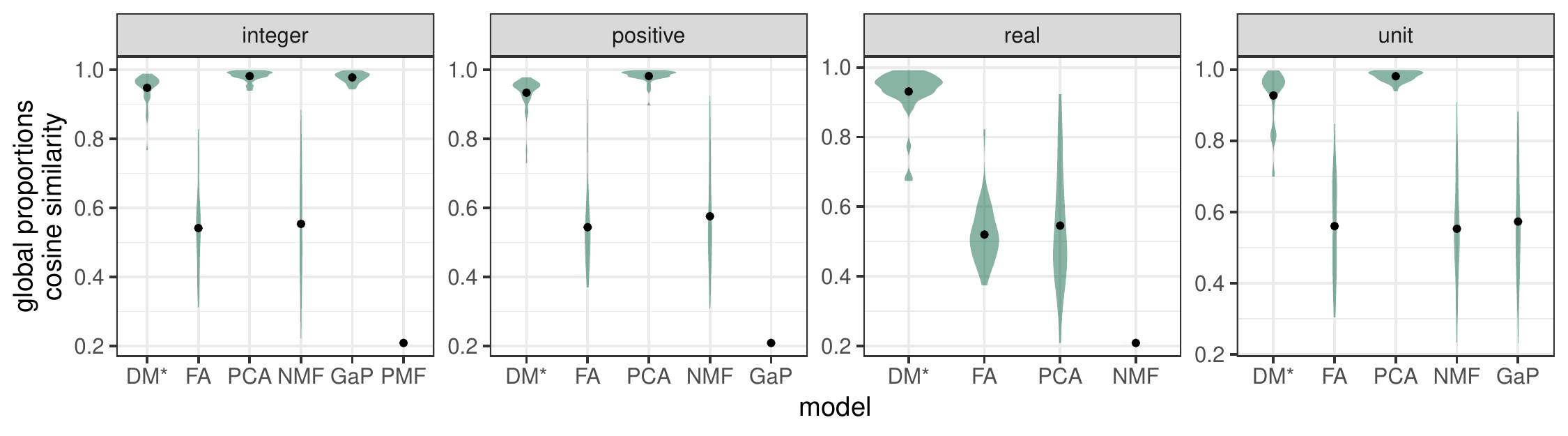}
\caption{Cosine similarity (\Cref{eq:cosine}, higher is better) of estimated global proportions $\hat{\boldsymbol{\beta}}$ and true simulated proportions $\boldsymbol{\beta}$, grouped by simulated data domain and averaged across all seeds and simulation procedures. DMs (starred) perform the best on data simulated using Gaussian distributions for $f$ (real domain) and performs equivalent to the best method in all other instances.}
\label{fig:beta_cosine}
\end{figure}

\begin{figure}
\centering
\includegraphics[width=\textwidth]{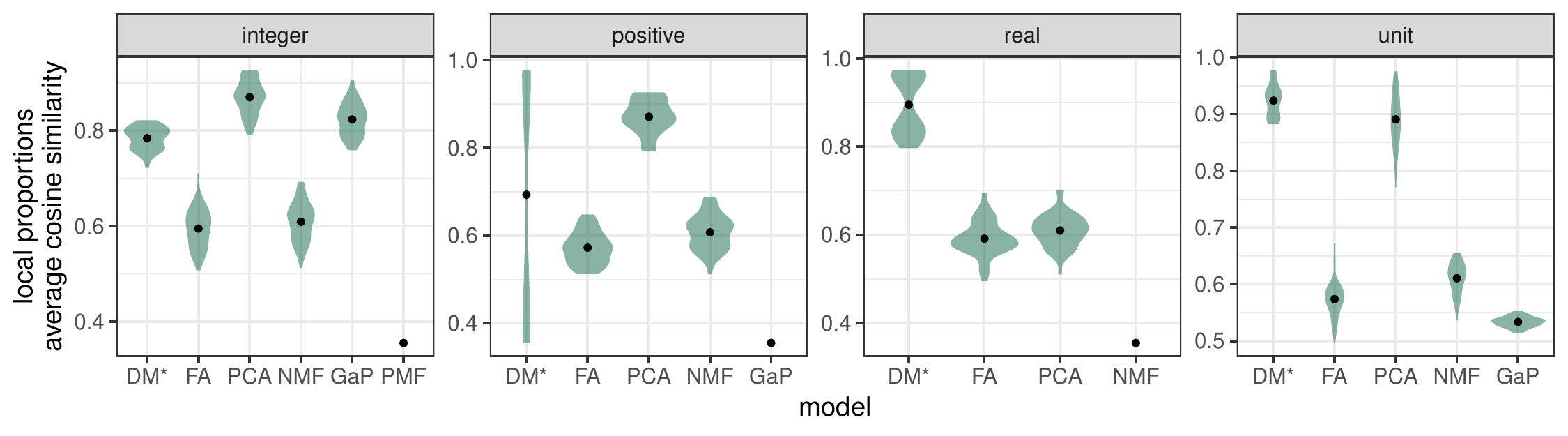}
\caption{Cosine similarity (\Cref{eq:cosine}, larger values indicate closer similarity) of estimated local proportions $\hat{\boldsymbol{\pi}}$ and true simulated proportions $\boldsymbol{\pi}$, averaged across $N$ observations in each simulated data set.  Results are grouped by simulated data domain and also averaged across all seeds and simulation procedures. DMs (starred) outperform comparison methods with data simulated using Gaussian (real domain) and beta (unit domain) distributions for $f$.  With data simulated using gamma $f$ (positive domain), DMs have high variance: one mode outperforms comparison methods and another mode under-performs.  With data simulated using Poisson $f$ (integer domain data), DMs perform well relative to comparison methods, FA and NMF in particular, but perform slightly worse than GaP and PCA (PCA performing the best overall).
}
\label{fig:pi_cosine}
\end{figure}

We also validated the model estimates of both global and local proportions ($\hat{\boldsymbol{\beta}}$ and $\hat{\boldsymbol{\pi}}$) to their known simulated values ($\boldsymbol{\beta}$ and $\boldsymbol{\pi}$).  To perform this comparison, we used the same fits of DM and comparison methods as described above and computed cosine similarity, or 
\begin{equation}
\mbox{cosine similarity}(\hat{\boldsymbol{v}}, \boldsymbol{v}) = \frac{\hat{\boldsymbol{v}} \cdot \boldsymbol{v} }{\vert\vert \hat{\boldsymbol{v}} \vert\vert~\vert\vert \boldsymbol{v} \vert\vert}.
\label{eq:cosine}
\end{equation}
For global proportions $\boldsymbol{\beta}$, we averaged cosine similarity across all factors $K$; for local proportions $\boldsymbol{\pi}$, we averaged cosine similarity across both observations $N$ and factors $K$.
In estimating global proportions $\boldsymbol{\beta}$, DMs  outperform all other methods with data simulated using a Gaussian distribution for $f$ (real domain); similarly, DMs perform close to the best comparison methods with all other simulated data (\Cref{fig:beta_cosine}).
For estimating local proportions $\boldsymbol{\pi}$, we found more nuanced results (\Cref{fig:pi_cosine}).  DMs outperform comparison methods with data simulated using Gaussian (real domain) and beta (unit domain) distributions for $f$.  For data simulated using gamma distributions for $f$ (positive domain), we found that DMs have high variance in the performance of the estimates with one mode outperforming the comparison methods and the other mode under-performing.  For data simulated using Poisson distributions for the link function $f$ (integer domain), DMs perform well relative to comparison methods, FA and NMF in particular, but do not yield the best estimates; in this domain, PCA and GaP slightly outperform DMs, with PCA yielding the best results.

\subsection{California Voting Data}
\label{sec:dat_vote}
As an example application, we explore the results of fitting a nonparametric deconvolution model (NDM) on voting data from the 2016 election in California.  These data can be modeled as count data and fit with a Poisson NDM, or as proportional (or unit-domain, or compositional) data and fit with a beta-distributed NDM.  In exploring both model types, we found that the model assuming count data identified voting cohorts well correlated with population size, and less correlated with shared voting behavior.  While discovering latent groups based on size may be desirable in some contexts, we opt for casting the data as proportional in order to study voting behavior.\looseness=-1

We fit a beta-distributed NDM on these proportional data with an initial $K=15$ factors, global concentration   parameter $\alpha_0=1$, local concentration parameter $\alpha=10$, local counts prior $\rho=100$, and other settings as the defaults in our release code.

\paragraph{Data description.}
We downloaded a data set of precinct-level votes on presidential candidates and propositions in the 2016 California Election, as provided by the LA Times.\footnote{More information about the data collection process may be found in the following LA Times Article: \url{http://www.latimes.com/projects/la-pol-ca-california-neighborhood-election-results/} and the data can be downloaded from\\ \url{https://github.com/datadesk/california-2016-election-precinct-maps}.}  These data included $N=24,568$ precincts with $M=37$ possible votes on candidates and propositions.

\paragraph{Model exploration.}
The first question we wanted to answer was: how many voting cohorts are there, and how do they vote?
Fitting an NDM to these data revealed ten voting cohorts; at face value, this makes sense because there are ten categories of party registration available---Democratic Party, Republican Party, American Independent Party, Libertarian Party, Green Party, Peace and Freedom Party, Reform Party, and Natural Law Party.
Registrants can also ``Decline to State'' or belong to Parties that are grouped together as ``Miscellaneous Parties.'' Registration data is available at \url{http://statewidedatabase.org/pub/data/G16/state/state_g16_registration_by_g16_rgprec.zip}.

\begin{figure}[p]
\centering
\includegraphics[width=0.938\textwidth]{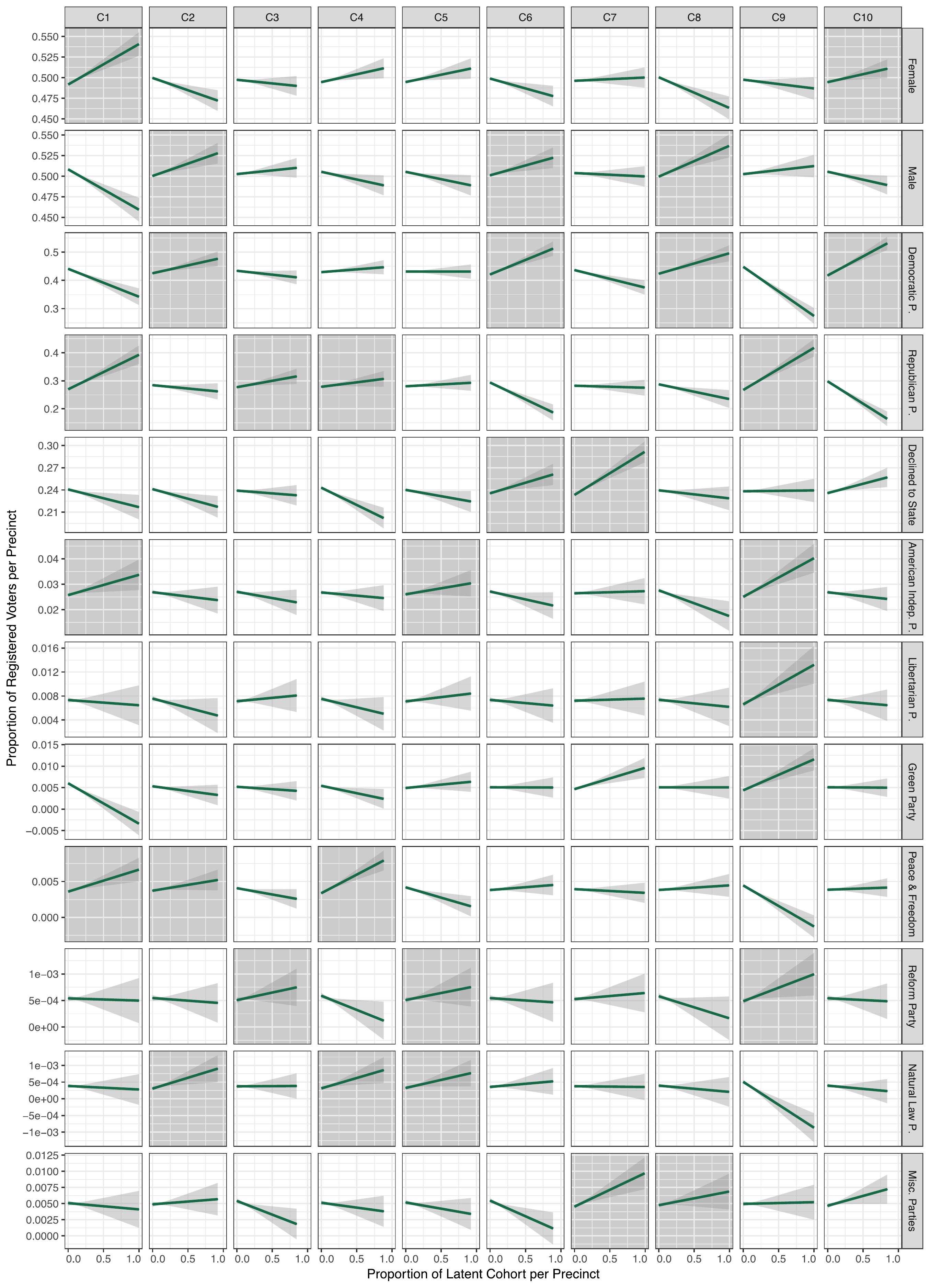}
\caption{Correlations between latent voting cohorts discovered with an NDM and actual voting registration information in the corresponding precinct.  Highlighted in grey are notable correlations; e.g., Cohort 10 correlates with female registered Democrats.}
\label{fig:demographics}
\end{figure}

However, we find that these cohorts do not precisely align with voter registration.  We found that the proportion of latent voter cohorts varies with the proportion of registered voters in each precinct (\Cref{fig:demographics}).  Cohort 1, for example, is more representative of female voters belonging to the Republican Party, but also correlates with American Independent Party and Peace and Freedom Party registrations.  Cohort 7 captures voters who decline to state a political party or belong to miscellaneous very small parties.

\begin{figure}[ht]
\centering
\includegraphics[width=\textwidth]{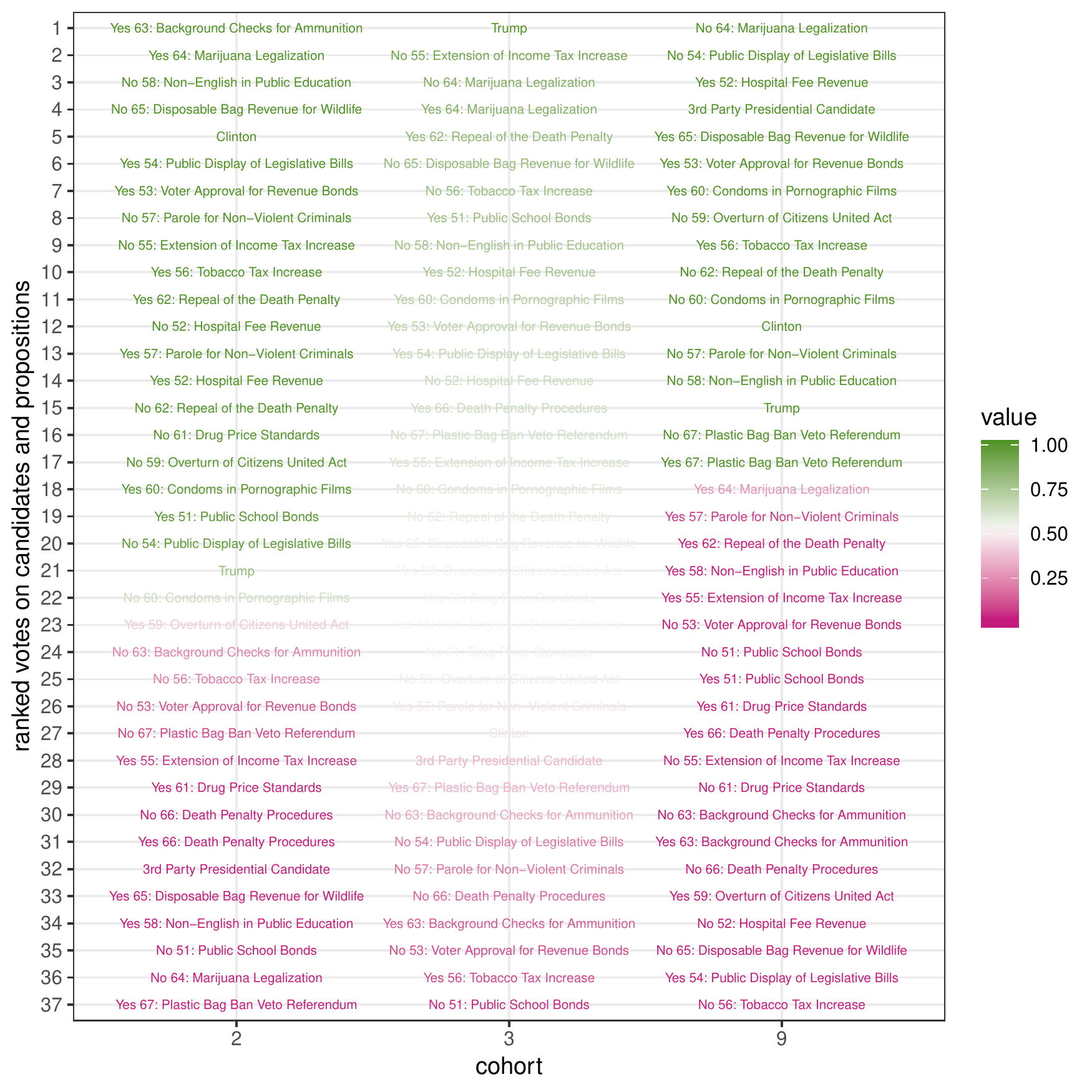}
\caption{Votes on candidates and propositions, grouped by cohort and ranked in order of the probability of an individual in that cohort casting the stated vote. Cohort 2 was strongly in favor of Proposition 63 on background check for ammunition, Cohort 3 was the most pro-Trump, and Cohort 9 was against the legalization of marijuana.}
\label{fig:topissues_3cohorts}
\end{figure}

Of greater interest is the voting patterns of these cohorts, and in particular the global voting patterns for individuals in three cohorts (\Cref{fig:topissues_3cohorts}) and the ranked issues for all cohorts (\Cref{fig:topissues_all}).  Cohort 2, which correlates with male Democrats in voter registration, was strongly in favor of Proposition 63 on background check for ammunition.  Cohort 3, which correlates with Republican Party and Reform Party registrations, was the most pro-Trump cohort.  Cohort 9, which correlates with several parties (Republican, American Independent, Libertarian, Green, and Reform), was against Proposition 64, which legalized marijuana.

\begin{figure}[p]
    \begin{subfigure}{\textwidth}
        \begin{subfigure}{.45\textwidth}
            \includegraphics[width=\textwidth]{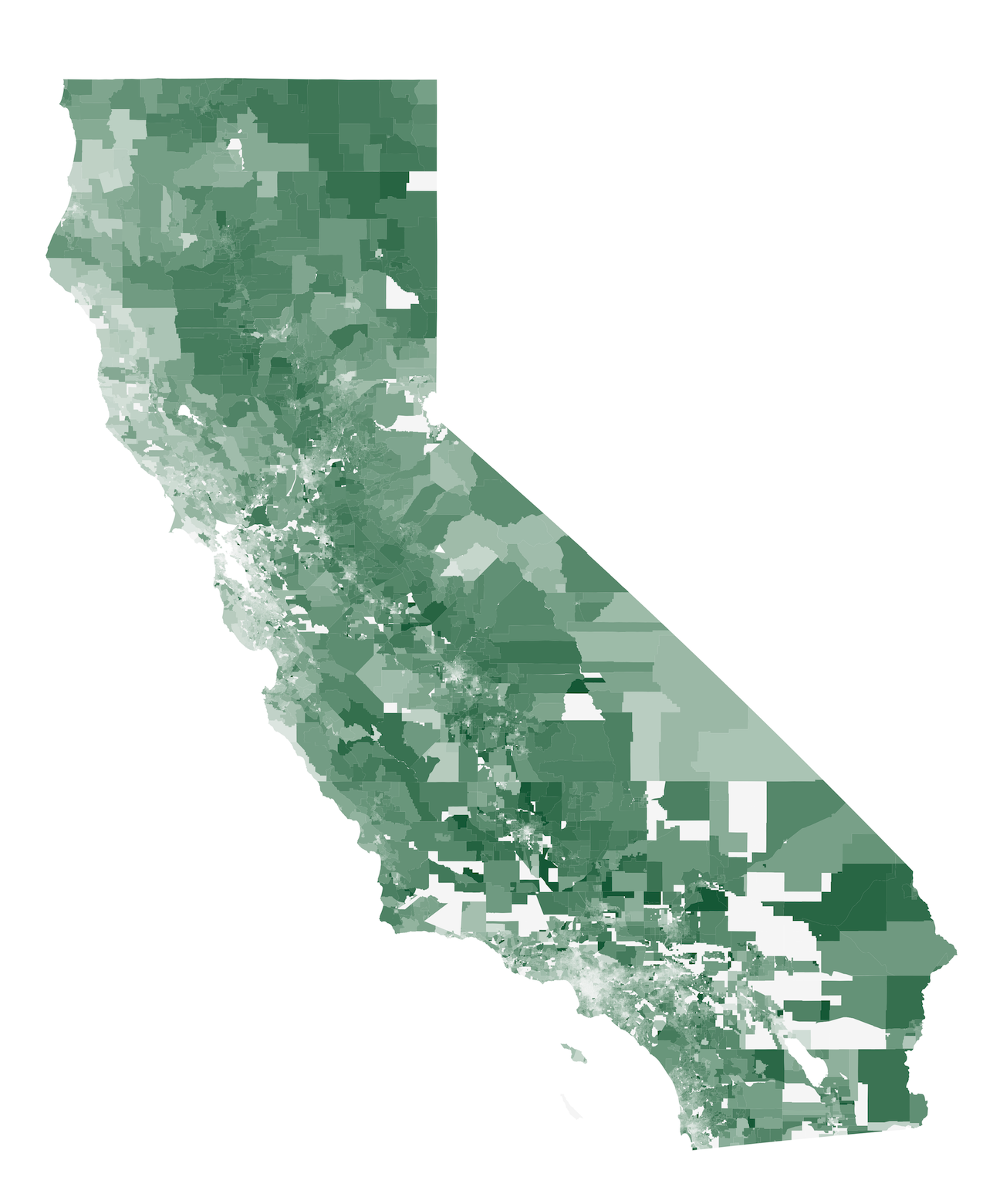}
            \caption*{\centering \footnotesize \textbf{Proportion of total population voting\newline for Trump (observed)}}
        \end{subfigure}
        \hfill
        \begin{subfigure}{.45\textwidth}
            \includegraphics[width=\textwidth]{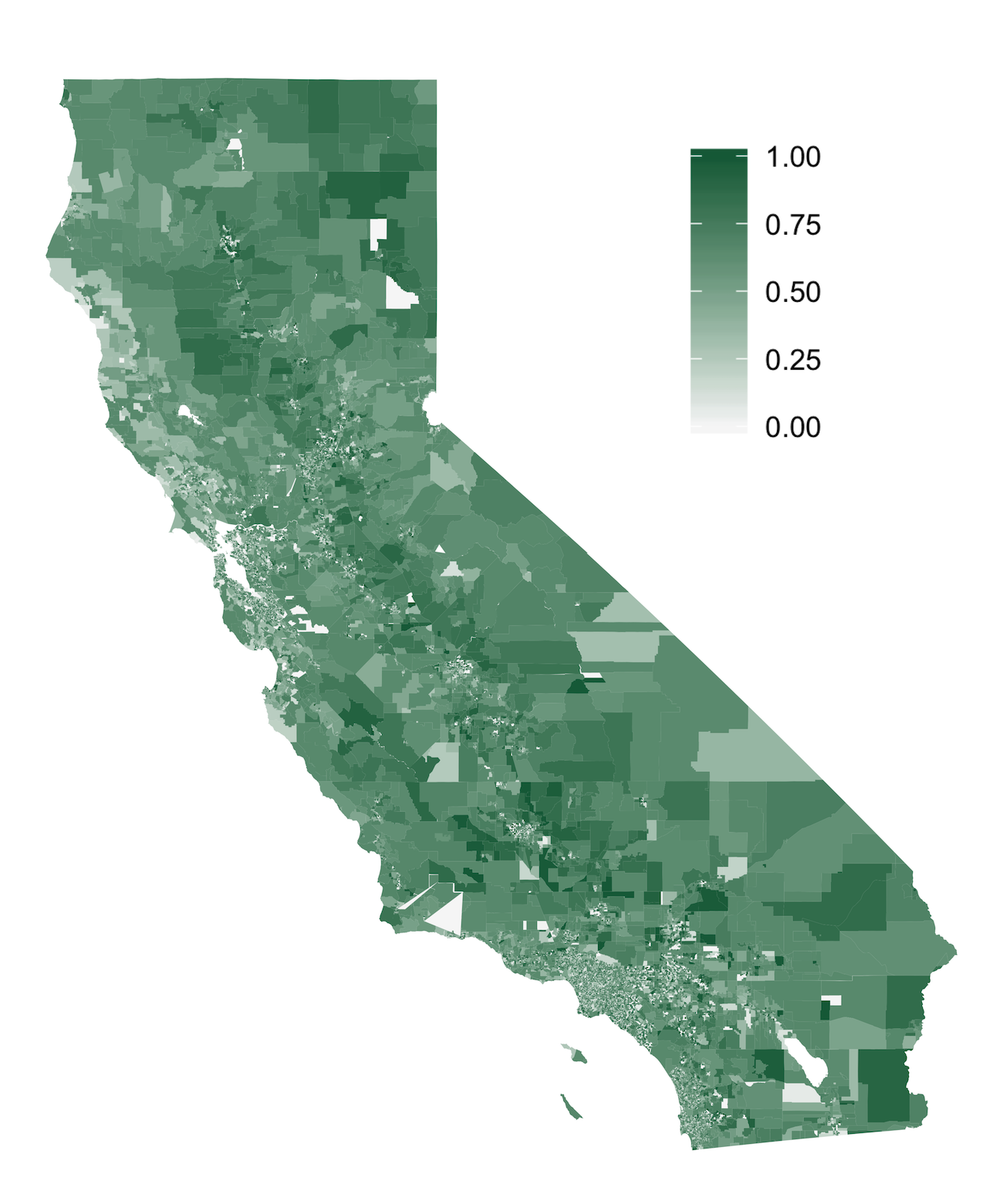}
            \caption*{\centering \footnotesize \textbf{Proportion of cohort 3 voting\newline for Trump (latent)}}
        \end{subfigure}
    \end{subfigure}
    \vspace{10px}
    \begin{subfigure}{\textwidth}
        \begin{subfigure}{.45\textwidth}
            \includegraphics[width=\textwidth]{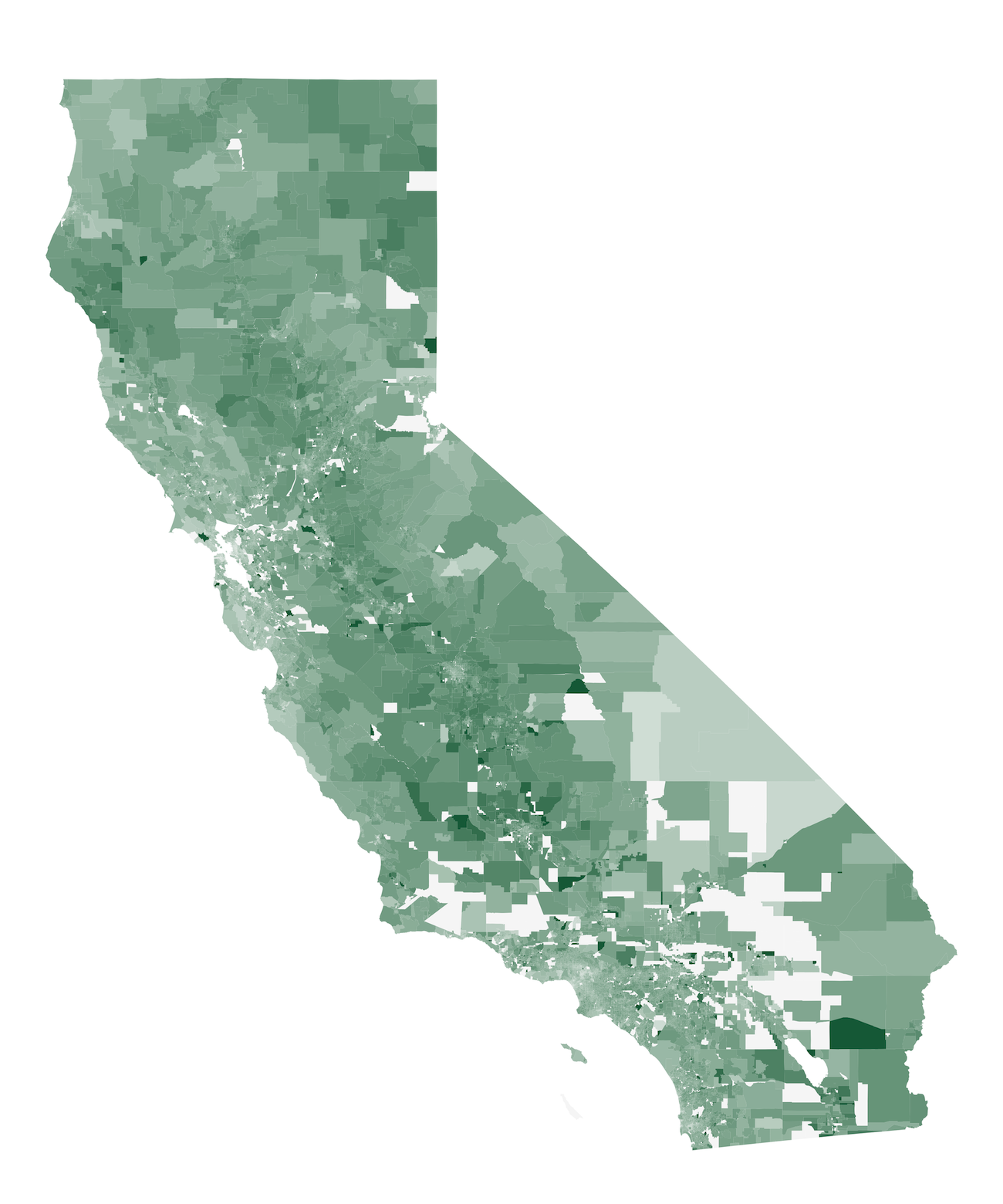}
            \caption*{\centering \footnotesize \textbf{Proportion of total population voting No on\newline Prop 64: Marijuana Legalization (observed)}}
        \end{subfigure}
        \hfill
        \begin{subfigure}{.45\textwidth}
            \includegraphics[width=\textwidth]{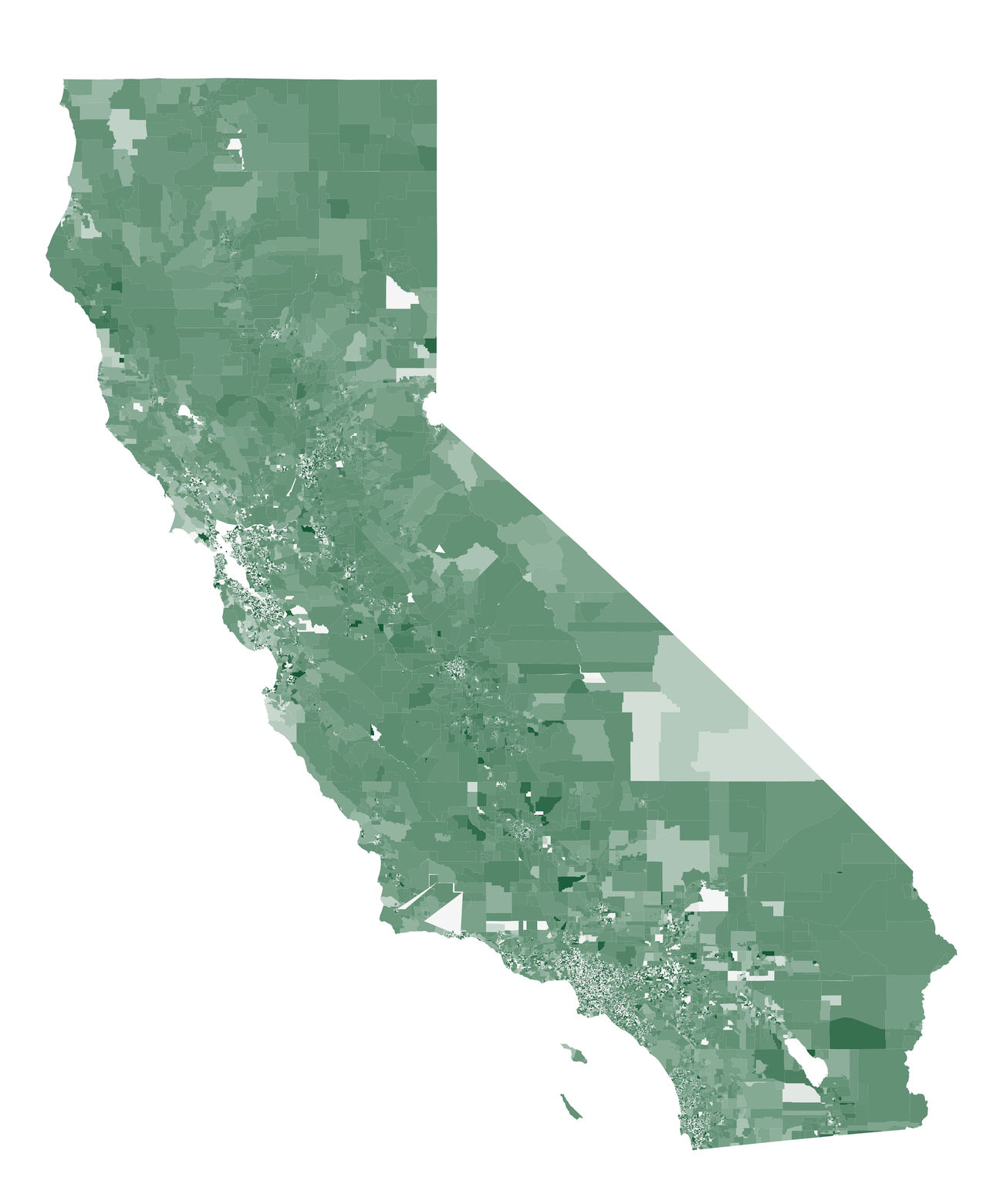}
            \caption*{\centering \footnotesize \textbf{Proportion of cohort 9 voting No on\newline Prop 64: Marijuana Legalization (latent)}}
        \end{subfigure}
    \end{subfigure}
    \vspace{10px}
    
    \caption{Observed total population (left) and modeled cohort (right) votes in the 2016 election.  Darker represents a higher proportion of votes and lighter a lower proportion for a candidate or issue.  Unlike existing models, NDMs captures the local fluctuations in latent voting cohorts, allowing us to make the maps on the right.}
    \label{fig:trump_maps}
\end{figure}

Global voting patterns like these can also be uncovered using an existing decomposition model; the unique power of deconvolution models comes from the ability to explore variance in the local features.  For example, we find that Cohort 8 shows the highest overall variance in voting patterns, and, conversely, that votes for third party candidates for president have the highest variance in each cohort.  By fitting an NDM to these voting data, we are able to estimate the local fluctuations in latent voting cohorts, allowing us to map out cohort-specific voting trends on candidates and propositions across precincts (e.g., \Cref{fig:trump_maps}), which allows us to identify cohorts in a specific precinct that differ from the global patterns of that cohort.  As an example, we find that, while Cohort 9 was generally against proposition 64 (legalized marijuana for use by adults 21 and over), that voters in this cohort from precincts in the Death Valley area were generally more in favor of this proposition.  This effect may be because Death Valley National Park has suffered from illegal marijuana cultivation sites; the National Park service has issued safety warnings on Marijuana Cultivation in
Death Valley National Park, e.g., \url{https://www.nps.gov/deva/planyourvisit/upload/DEVA-Marijuana-Safety.pdf}. Legalization of marijuana would likely diminish these occurrences.  Identifying anomalies in cohort voting behavior such as these with NDMs could be a step toward discovering new ways to identify individuals to approach for candidates and issues.\looseness=-1

\section{Discussion}
\label{sec:conclusion}
Our nonparametric deconvolution model (NDM) addresses the problem of modeling collections of convolved data points.  Unlike decomposition and admixture models, which model latent factor feature distributions as the same for each observation, our proposed deconvolution model family captures how the feature distributions vary for each observation, allowing us to explicitly model variation in latent factors in the context of each observation.  This general model family can be applied to data with various domains by choosing an appropriate distribution $f$ and link function $g$.  

Our contributions include the specification of the deconvolution model family (both parametric and nonparametric), developing an inference framework for this family, and releasing source code for our inference algorithm.  We explore the performance of NDMs empirically on simulated and California voting data.  We found that modeling local factor features leads to better estimates of latent variables (factor features and proportions).  NDMs also provides a novel framework for exploring data, as demonstrated by our study of NDM results on 2016 California voting data.

There are many avenues for future work on deconvolution models.  For example, the inference framework we provided for the general NDM family is sensitive to the learning rate on local factor features $\boldsymbol{\bar{x}}$; techniques can be developed to reduce this sensitivity.  The speed of the inference algorithm can also be improved.  In terms of modeling improvements, a natural extension of this model family is to include covariate information.  For example, with voting data, census or registration information can be incorporated into the model directly.  In some applications, this covariate data is available at finer scales than the data we wish to deconvolve (in the earth sciences, this problem is commonly known as downscaling); this information can be used to provide better estimates of local feature distributions.

\acks{AJBC was supported by an appointment to the Intelligence Community Postdoctoral Research Fellowship Program at Princeton University, administered by Oak Ridge Institute for Science and Education through an interagency agreement between the U.S. Department of Energy and the Office of the Director of National Intelligence. BEE was funded by NIH R01 MH101822, NIH R01 HL133218, a Sloan Faculty Fellowship, and an NSF CAREER AWD1005627.}

\bibliography{generalized_ndm.bib}

\appendix
\crefalias{Subsection}{appsec}
\crefalias{Section}{appsec}

\section{Inference Algorithm Details}
This appendix outlines minutiae relevant to inference for replicability.  Readers are also invited to explore our open-source implementation of the algorithm (\url{https://github.com/ajbc/ndm}) for questions that are not addressed here, with the caveats that the software is academic and not developed with speed or industry-style coding standards in mind.

\subsection{Partial log joint probabilities containing only relevant terms}
\label{app:partiallogjoints}
As described in \Cref{sec:parametric_inference}, black box variational inference for some parameter $z$ requires the log probability of all terms containing the hidden parameter of interest, or $\log p^{z}$.  For example, the log probability for local features $\boldsymbol{\bar{x}}_{n,k}$ is defined as follows (also shown in \Cref{eq:partiallog}):
\begin{equation}
    \log p^{\bar{x}}_{n,k}( \boldsymbol{y}, \boldsymbol{\Sigma}, \boldsymbol{\beta}, \boldsymbol{\pi}, \boldsymbol{\mu}, \boldsymbol{\bar{x}}, \boldsymbol{P}) \triangleq \log p(\boldsymbol{\bar{x}}_{n,k} \g \boldsymbol{\mu}_k, \boldsymbol{\Sigma}_k, \boldsymbol{\pi}_n, P_n) + \log p(y_{n} \g \boldsymbol{\bar{x}}_{n}, \boldsymbol{\pi}_n).
    \label{eq:partiallog_x}
\end{equation}

We now write out the other partial log joint probabilities containing only relevant terms.  For local counts $\boldsymbol{P}$, we have
\begin{equation}
    \log p^{P}_{n}( \boldsymbol{y}, \boldsymbol{\Sigma}, \boldsymbol{\beta}, \boldsymbol{\pi}, \boldsymbol{\mu}, \boldsymbol{\bar{x}}, \boldsymbol{P}) \triangleq \log p(\boldsymbol{\bar{x}}_{n,k} \g \boldsymbol{\mu}_k, \boldsymbol{\Sigma}_k, \boldsymbol{\pi}_n, P_n) + \log p(P_n \g \rho).
    \label{eq:partiallog_p}
\end{equation}

For local proportions $\boldsymbol{\pi}$, we use the following partial log joint.
\begin{equation}
    \log p^{\pi}_{n}( \boldsymbol{y}, \boldsymbol{\Sigma}, \boldsymbol{\beta}, \boldsymbol{\pi}, \boldsymbol{\mu}, \boldsymbol{\bar{x}}, \boldsymbol{P}) \triangleq 
    \sum_{k=1}^{K}\log p(\boldsymbol{\bar{x}}_{n,k} \g \boldsymbol{\mu}_k, \boldsymbol{\Sigma}_k, \boldsymbol{\pi}_n, P_n) +
    \log p(y_{n} \g \boldsymbol{\bar{x}}_{n}, \boldsymbol{\pi}_n) +
    \log p(\pi_{n} \g \alpha, \boldsymbol{\beta}).
    \label{eq:partiallog_pi}
\end{equation}

For global proportions $\boldsymbol{\beta}$, we have
\begin{equation}
    \log p^{\beta}( \boldsymbol{y}, \boldsymbol{\Sigma}, \boldsymbol{\beta}, \boldsymbol{\pi}, \boldsymbol{\mu}, \boldsymbol{\bar{x}}, \boldsymbol{P}) \triangleq 
    \log p(\boldsymbol{\beta} \g \boldsymbol{\alpha}_0) +
    \sum_{n=1}^N\log p(\pi_{n} \g \alpha, \boldsymbol{\beta}).
    \label{eq:partiallog_beta_param}
\end{equation}

\subsection{Gradients}
\label{app:gradients}
Black box variational inference also requires gradients of the approximating family of distributions $q$ with respect to the variational parameters of interest $\boldsymbol{\lambda}[z]$, or $\nabla_{\boldsymbol{\lambda}[z]} \log q(z \g \boldsymbol{\lambda}[z])$.  The gradients used in our algorithm are as follows.

The gradients of the normal distribution with respect to mean $\mu$ and variance $\sigma$, respectively, are
\begin{equation}
    \nabla_{\mu} \mathcal{N}(x \g \mu, \sigma) = (x - \mu) / \sigma^2,
\end{equation}
and
\begin{equation}
    \nabla_{\sigma} \mathcal{N}(x \g \mu, \sigma) = -\frac{1}{\sigma} + \frac{(x - \mu)^2}{\sigma^3}.
\end{equation}

The gradient of the Poisson with respect to its rate $\lambda$ is
\begin{equation}
    \nabla_{\lambda} \mbox{Poisson}(x \g \lambda) = \frac{x}{\lambda} - 1.
\end{equation}

The gradient of the Dirichlet distribution with respect to its concentration parameters $\boldsymbol{\alpha}$ is
\begin{equation}
    \nabla_{\boldsymbol{\alpha}} \mbox{Dirichlet}(x \g \boldsymbol{\alpha}) = 
    \log(x) + \psi\left(\sum_{k=1}^K \alpha_k\right) - \psi(\boldsymbol{\alpha}),
\end{equation}
where $\psi$ is the digamma function.

\subsection{Learning rates.}
\label{app:learningrates}
We use the following construction for learning rates: at iteration $t$ the learning rate is
\begin{equation}
    \rho_t = (t + d)^r,
\end{equation}
where delay $d \ge 0$ down-weights early iterations and rate $r \in (0.5,1]$ impacts how quickly old information is forgotten.  We generally use the following learning rates.\footnote{We use slightly different learning rates for local features $\boldsymbol{\bar{x}}$ when the distribution $f$ for the observations $\boldsymbol{y}$ is beta; in this case, the delay $d$ on the location is set to $2^{10}$.}

\begin{center}
\begin{tabular}{rrcc}
    \toprule
    \textbf{hidden variable} & \textbf{variational parameter} & $d$ & $r$\\
    \midrule
    global proportions $\boldsymbol{\beta}$ & 
        concentration $\boldsymbol{\lambda}[\beta(\alpha)]$& $2^4$ & -0.5 \\
    local proportions $\boldsymbol{\pi}$ & 
        concentration $\boldsymbol{\lambda}[\pi(\alpha)]$ & $2^{10}$ & -0.8 \\
    local features $\boldsymbol{\bar{x}}$ & 
        location $\boldsymbol{\lambda}[\bar{x}(\mu)]$ & $2^{20}$ & -0.8 \\
    local features $\boldsymbol{\bar{x}}$ &
        scale $\boldsymbol{\lambda}[\bar{x}(\sigma)]$ & $2^{20}$ & -0.8\\
    local counts $\boldsymbol{P}$ &
        rate $\boldsymbol{\lambda}[P(\mu)]$ & $2^5$ & -0.7\\
    \bottomrule
\end{tabular}
\end{center}

Even with the extensions to control gradient variance (\Cref{sec:parametric_inference}), we find that the variances for estimates of local variables (proportions $\boldsymbol{\pi}$, features $\boldsymbol{\bar{x}}$, and counts $\boldsymbol{P}$) are particularly high.  This is somewhat unsurprising: we would expect that the true posterior variances of these variables are high, since we are estimating these parameters from aggregated data $\boldsymbol{y}$.
Because of the variances in local parameter estimates, we find that our method is somewhat sensitive to the learning rates $\rho$, which need to be set carefully.  The learning rate for local features $\boldsymbol{\bar{x}}$ is especially important, as this is the aspect of the model that can easily overfit.  Reducing the sensitivity to the learning rates is an avenue for future work.

\subsection{Initialization}
\label{app:initialize}
To initialize the variational parameters $\boldsymbol{\lambda}$, we begin by adding a small amount of random noise to the data and then fit a fuzzy $K$-means model.  The resulting fuzzy $K$-means labels for each observation are scaled and used to initialize the concentration variational parameters for the local proportions $\boldsymbol{\pi}$; the average of these labels is used to initialized the variational parameters for the global proportions $\boldsymbol{\beta}$ (also scaled).  The fuzzy $K$-means centroids are used to initialize the mean variational parameters for global factor centers $\boldsymbol{\mu}$ and also the local factor centers $\bar{\boldsymbol{x}}$.   The global factor covariances $\boldsymbol{\Sigma}$ are initialized as matrices with the variances of each observed feature along the diagonal.

In the nonparametric variant, the initialization is adjusted slightly.  In particular, the local and global proportions ($\boldsymbol{\pi}$ and $\boldsymbol{\beta}$, respectively) require initial vales for the $K+1$ location, which are set to relatively small numbers.  The factor features (local and global) must also be initialized; the global mean variational parameter $\boldsymbol{\lambda}[\mu_{k+1}[\mu]]$ for this last catch-all factor is set to the mean values of the data (appropriately transformed with $g^{-1}$) and the local mean variational parameter  $\boldsymbol{\lambda}[\bar{x}_{n,k+1}[\mu]]$ is set to residual, or the difference between the observed values $\boldsymbol{y_n}$ and the reconstruction of that observation with the first $K$ values of $\boldsymbol{\bar{x}}_n$ and $\boldsymbol{\pi}_n$.

\subsection{Convergence Criteria}
\label{app:convergence}

In \Cref{sec:inference}, we describe the termination criteria for our inference algorithm as being when the ``change in ELBO $< \delta$;'' this is not the full representation of the convergence criteria and we will provide further details here.

Since our inference procedure has some stochastic elements, there may be some fluctuations in the ELBO.  Thus, we have a ``convergence counter'' that increments each time the relative change in the ELBO is sufficiently small ($< \delta=0.0001$).  When the counter reaches above three consecutive iterations of meeting the convergence criteria, we terminate the inference algorithm, provided the minimum number of iterations has been met.  Alternatively, the inference algorithm can just be run for a fixed number of iterations.

In the nonparametric setting, there are a few additional complications: we assess batch convergence prior to the split or merge procedures and we want to ensure that the parameters are not deemed converged too soon after a split or merge procedure.  For the former, we simply move to splitting or merging after a single instance of a low relative change in the ELBO ($< \delta=0.0001$) or once a batch iteration maximum has been reached.  For the latter, we reset the global convergence counter on a split or merge.

\section{Simulation Procedures}
\label{sec:app_sim}

We simulate data in five distinct ways.  Each procedure requires setting a fixed number of factors $K$, the number observations $N$, the number of features $M$, and the domain of the observations.
The domain may be real-valued, positive real numbers, positive integers, or in the unit interval $[0,1]$. 

To allow for these simulation procedures to generate data in different domains, we define a distribution $f$ for each of the domains.  If the specified domain is real values, $f$ is a normal distribution with mean and scale parameters, or 
\begin{equation}
    f(x \g \mu, \sigma) = \mathcal{N}(x \g \mu, \sigma) = \frac{1}{\sqrt{2\pi \sigma^2}} \exp\left(-\frac{(x-\mu)^2}{2\sigma^2}\right).
\end{equation}
If the domain is positive real numbers, the first parameter is transformed with the soft-plus function ($s(\mu)$) and we use a Gamma distribution with mean and shape parameters, or
\begin{equation}
    f(x \g \mu, \sigma) = \mbox{Gamma}^*(x \g s(\mu), \sigma);
\end{equation}
to put this in term of the more typical shape-scale Gamma parameterization, we have shape $a=(s(\mu)/\sigma)^2$,  scale $b=\sigma^2 / s(\mu)$, and $\mbox{Gamma}(x \g a, b) = \frac{x^{(a-1)}\exp(-x/b)}{\Gamma(a)b^a}$.

If the domain is positive integers, then we use a Poisson distribution; the first parameter is transformed with the soft-plus function $s(\mu)$ and the second parameter is ignored, or
\begin{equation}
    f(x \g \mu, \sigma) = \mbox{Poisson}(x \g \mbox{s}(\mu)) = \frac{\mbox{s}(\mu)^x \exp\left(-\mbox{s}(\mu)\right)}{x!}.
\end{equation}
If the domain is the unit interval $[0,1]$, then we use an atypically specified Beta distribution with the first parameter being transformed with a sigmoid function $S(\mu)$, or
\begin{equation}
    f(x \g \mu, \sigma) = \mbox{Beta}^*(x \g S(\mu), \sigma);
\end{equation}
to put this in term of the typical Beta parameterization, we have shape $a=\left(\frac{(1.0-S(\mu)}{\sigma^2} - \frac{1}{S(\mu)}\right) S(\mu)^2$,  shape $b=a \left(1/S(\mu) - 1\right)$, and $\mbox{Beta}(x \g a, b) = \frac{x^{(a-1)}(1-x)^{(b-1)}}{B(a,b)}$.   Because of this parameterization, we use a link function $g(x) = 1\times10^{-6} + (1-2\times10^{-6}) / (1 + \exp(-10(x -0.5)))$ with the logistic function parameters chosen to avoid parameter errors for extreme values, as well as with a logistic growth rate that generates interesting data for deconvolution (too shallow a rate leads to only extreme values for $\mu$, which makes $x$ too easy to predict).

When not explicitly defined by the simulation procedure, we define $\boldsymbol{\bar{x}}_{n,k} = \frac{1}{P_n} \sum_{p=1}^{P_n} \boldsymbol{1}[z_{n,p}=k] ~x_{n,p,m}$. We similarly define the local proportions $\boldsymbol{\pi}_{n,k}$ based on the number of particles assigned to observation $n$ and factor $k$ divided by the total number of particles assigned to observation $n$, or $\boldsymbol{\pi}_{n,k} = \frac{\sum_{p} \boldsymbol{1}[z_p = k ~\&~ a_p = n]}{\sum_{p} \boldsymbol{1}[a_p = n]}$.

\subsection{Simulation Procedure 1}
\label{app:sim1}
\begin{itemize}
    \item Draw global factor proportions $\boldsymbol{\beta} \sim \mbox{Dirichlet}(\boldsymbol{\alpha_0})$
    \item For factor $k=1,\dots,K$:
    \begin{itemize}
        \item Draw global factor feature means $\boldsymbol{\mu}_k \sim \mathcal{N}(\mu_0, \sigma)$
        \item Draw global factor feature covariances $\boldsymbol{\Sigma}_k \sim \mathcal{W}^{-1}(\boldsymbol{\Psi}, \nu)$
    \end{itemize}
    \item For feature $m=1,\dots,M$: (optional, depending on domain specification)
    \begin{itemize}
        \item Draw scale $\sigma_m \sim \mbox{Gamma}^{-1}(a, b)$
    \end{itemize}
    \item For observation $n=1,\dots,N$:
    \begin{itemize}
        \item Draw local proportions $\boldsymbol{\pi}_n \sim \mbox{Dirichlet}( \boldsymbol{\beta}*\alpha)$
        \item For factor $k=1,\dots,K$:
        \begin{itemize}
            \item Draw local factor feature means $\boldsymbol{\bar{x}}_k \sim \mathcal{N}(\boldsymbol{\mu}_k, \boldsymbol{\Sigma}_k)$
        \end{itemize}
        \item Draw counts $P_n \sim \mbox{Poisson}(\rho)$
        \item For $p=1,\dots,P_n$:
        \begin{itemize}
            \item Draw assignment $z_{n,p} \sim \mbox{Categorical}(\boldsymbol{\pi}_n)$
            \item Draw local features $\boldsymbol{x}_{n,p} \sim \mathcal{N}_M(\boldsymbol{\bar{x}}_{z_{n,p}}, \boldsymbol{\Sigma}_{z_{n,p}}\cdot10^{-6})$
        \end{itemize}
        \item For feature $m=1,\dots,M$:
        \begin{itemize}
            \item Draw observations $y_{n,m} \sim f\left(\frac{1}{P_n}\sum_{p=1}^{P_n} x_{n,p,m}, \sigma_m \right)$ 
        \end{itemize}
    \end{itemize}
\end{itemize}

\subsection{Simulation Procedure 2}
\label{app:sim2}
\begin{itemize}
    \item Draw global factor proportions $\boldsymbol{\beta} \sim \mbox{Dirichlet}(\boldsymbol{\alpha_0})$
    \item For factor $k=1,\dots,K$:
    \begin{itemize}
        \item Draw global factor feature means $\boldsymbol{\mu}_k \sim \mathcal{N}(\mu_0, \sigma)$
        \item Draw global factor feature covariances $\boldsymbol{\Sigma}_k \sim \mathcal{W}^{-1}(\boldsymbol{\Psi}, \nu)$
    \end{itemize}
    \item For feature $m=1,\dots,M$: (optional, depending on domain specification)
    \begin{itemize}
        \item Draw scale $\sigma_m \sim \mbox{Gamma}^{-1}(a, b)$
    \end{itemize}
    \item For observation $n=1,\dots,N$:
    \begin{itemize}
        \item Draw local proportions $\boldsymbol{\pi}_n \sim \mbox{Dirichlet}( \boldsymbol{\beta}*\alpha)$
        \item Draw counts $P_n \sim \mbox{Poisson}(\rho)$
        \item For $p=1,\dots,P_n$:
        \begin{itemize}
            \item Draw assignment $z_{n,p} \sim \mbox{Categorical}(\boldsymbol{\pi}_n)$
            \item Draw local features $\boldsymbol{x}_{n,p} \sim \mathcal{N}_M(\boldsymbol{\mu}_{z_{n,p}}, \boldsymbol{\Sigma}_{z_{n,p}})$
        \end{itemize}
        \item For feature $m=1,\dots,M$:
        \begin{itemize}
            \item Draw observations $y_{n,m} \sim f\left(\frac{1}{P_n}\sum_{p=1}^{P_n} x_{n,p,m}, \sigma_m \right)$ 
        \end{itemize}
    \end{itemize}
\end{itemize}

\subsection{Simulation Procedure 3}
\label{app:sim3}
\begin{itemize}
    \item Draw global factor proportions $\boldsymbol{\beta} \sim \mbox{Dirichlet}(\boldsymbol{\alpha_0})$
    \item For factor $k=1,\dots,K$:
    \begin{itemize}
        \item Draw global factor feature means $\boldsymbol{\mu}_k \sim \mathcal{N}(\mu_0, \sigma)$
        \item Draw global factor feature covariances $\boldsymbol{\Sigma}_k \sim \mathcal{W}^{-1}(\boldsymbol{\Psi}, \nu)$
    \end{itemize}
    \item For feature $m=1,\dots,M$: (optional, depending on domain specification)
    \begin{itemize}
        \item Draw scale $\sigma_m \sim \mbox{Gamma}^{-1}(a, b)$
    \end{itemize}
    \item For observation $n=1,\dots,N$:
    \begin{itemize}
        \item Draw local proportions $\boldsymbol{\pi}_n \sim \mbox{Dirichlet}( \boldsymbol{\beta}*\alpha)$
        \item For factor $k=1,\dots,K$:
        \begin{itemize}
            \item Draw local factor feature means $\boldsymbol{\bar{x}}_k \sim \mathcal{N}(\boldsymbol{\mu}_k, \boldsymbol{\Sigma}_k)$
        \end{itemize}
        \item Draw counts $P_n \sim \mbox{Poisson}(\rho)$
        \item For $p=1,\dots,P_n$:
        \begin{itemize}
            \item Draw assignment $z_{n,p} \sim \mbox{Categorical}(\boldsymbol{\pi}_n)$
            \item Draw local features $\boldsymbol{x}_{n,p} \sim f(\boldsymbol{\bar{x}}_{z_{n,p}}, \sigma_m)$
        \end{itemize}
        \item For feature $m=1,\dots,M$:
        \begin{itemize}
            \item Set observations $y_{n,m} = \frac{1}{P_n}\sum_{p=1}^{P_n} x_{n,p,m}$ 
        \end{itemize}
    \end{itemize}
\end{itemize}

\subsection{Simulation Procedure 4}
\label{app:sim4}
\begin{itemize}
    \item Draw global factor proportions $\boldsymbol{\beta} \sim \mbox{Dirichlet}(\boldsymbol{\alpha_0})$
    \item For factor $k=1,\dots,K$:
    \begin{itemize}
        \item Draw global factor feature means $\boldsymbol{\mu}_k \sim \mathcal{N}(\mu_0, \sigma)$ 
        \item Draw global factor feature covariances $\boldsymbol{\Sigma}_k \sim \mathcal{W}^{-1}(\boldsymbol{\Psi}, \nu)$
        \item Draw a number of modes $S_k \sim \mbox{Poisson}(5)$; \hspace{10pt} (note: keep drawing until $S_k>0$)
        \item For mode $s=1,\dots,S_k$:
        \begin{itemize}
            \item Draw mode features $\boldsymbol{\mu}_k[s] \sim \mathcal{N}_M(\boldsymbol{\mu}_{k}, \boldsymbol{\Sigma}_{k})$
        \end{itemize}
        
    \end{itemize}
    \item For feature $m=1,\dots,M$: (optional, depending on domain specification)
    \begin{itemize}
        \item Draw scale $\sigma_m \sim \mbox{Gamma}^{-1}(a, b)$
    \end{itemize}
    \item For observation $n=1,\dots,N$:
    \begin{itemize}
        
        \item Draw $\boldsymbol{\pi}_n \sim \mbox{Dirichlet}( \boldsymbol{\beta})$
        \item Draw $P_n \sim \mbox{Poisson}(\rho)$
        \item For $p=1,\dots,P_n$:
        \begin{itemize}
            \item Draw assignment $z_{n,p} \sim \mbox{Categorical}(\boldsymbol{\pi}_n)$
            \item Draw mode $s_{n,p} \sim \mbox{Categorical}(\boldsymbol{\gamma}_{z_{n,p}})$
            \item For feature $m=1,\dots,M$:
            \begin{itemize}
                \item Draw local features $x_{n,p,m} \sim f\left(\boldsymbol{\mu}_{z_{n,p},m}[s_{n,p}], \sigma_m \right)$
            \end{itemize}
        \end{itemize}
        \item For feature $m=1,\dots,M$:
        \begin{itemize}
            \item Set observations $y_{n,m} = \frac{1}{P_n}\sum_{p=1}^{P_n} x_{n,p,m}$ 
        \end{itemize}
    \end{itemize}
\end{itemize}

\section{Additional Empirical Results}
\label{sec:app_detailed_results}
\begin{sidewaysfigure}[ht]
\centering
\includegraphics[width=\textwidth]{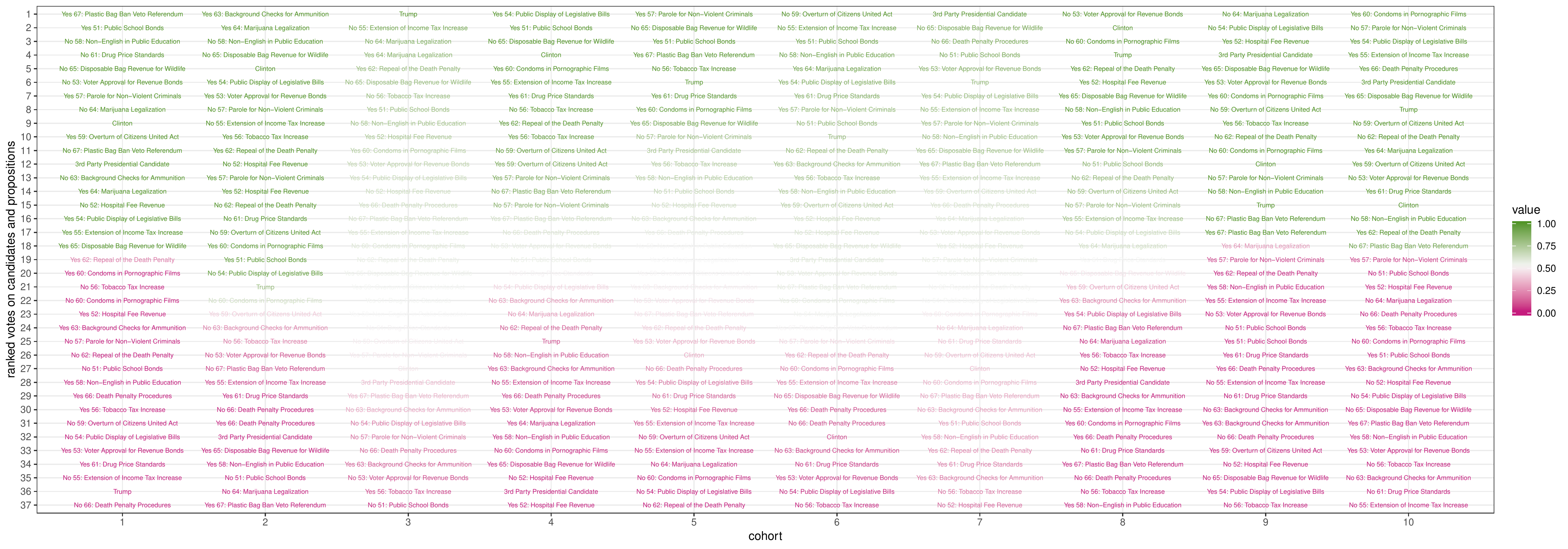}
\caption{Votes on candidates and propositions, grouped by cohort and ranked in order of the probability of an individual in that cohort casting the stated vote.  For example, Cohort 2 was strongly in favor of Proposition 63 on background check for ammunition, Cohort 3 was the most pro-Trump, and Cohort 9 was against the legalization of marijuana.}
\label{fig:topissues_all}
\end{sidewaysfigure}

Here we expand \Cref{fig:topissues_3cohorts} to include all cohorts; the result is \Cref{fig:topissues_all}.

\end{document}